\documentclass[10pt,twocolumn,letterpaper]{article}

\usepackage[pagenumbers]{cvpr} 

\usepackage{multirow}
\usepackage{graphicx}
\usepackage{tabularx}  
\usepackage{booktabs}  
\usepackage{amsmath}   
\usepackage{colortbl}    
\usepackage{array}       
\usepackage{hhline}
\usepackage{arydshln}

\usepackage[dvipsnames]{xcolor}

\usepackage{amssymb}
\usepackage{pifont}
\definecolor{cvprblue}{rgb}{0.21,0.49,0.74}
\usepackage[pagebackref,breaklinks,colorlinks,allcolors=cvprblue]{hyperref}

\newlength{\tablewidth}
\def\paperID{4242} 
\def\confName{CVPR}
\def\confYear{2025}

\newcommand{\ourmethod}{\text{DocVLM} }
\newcommand{\refwithdefault}[2]{%
  \ifcsname r@#1\endcsname
    \ref{#1}%
  \else
    #2%
  \fi
}
\definecolor{lightgray}{gray}{0.9}  

\title{
\vspace{-0.8cm}
\hspace{-1.8em}\includegraphics[width=1.5cm]{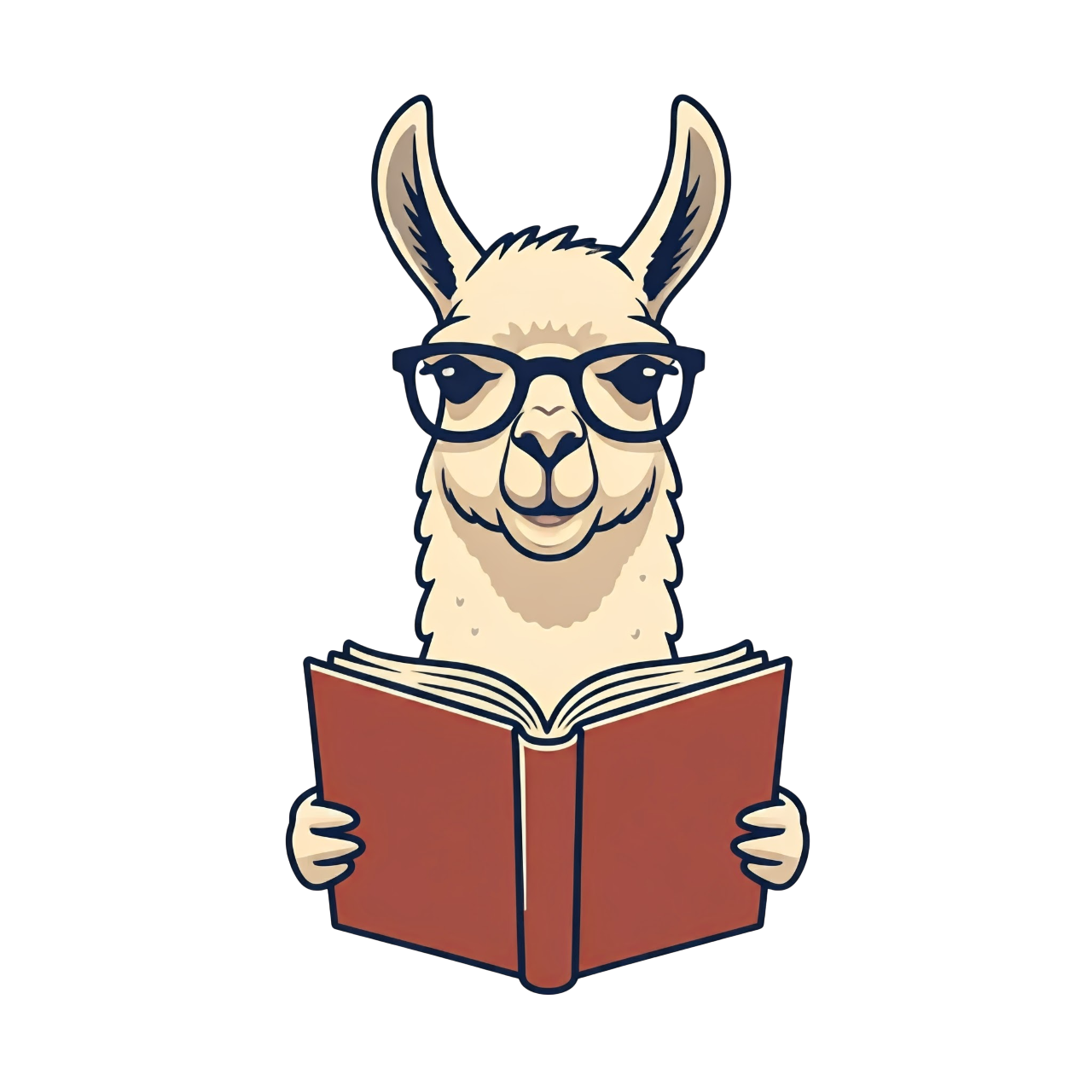}  
\hspace{0.2em}
\raisebox{0.55cm}{\textbf{
DocVLM: Make Your VLM an Efficient Reader
}}
\vspace{-0.5cm}
}

\def\maketitlesupplementary
   {
   \newpage
       \twocolumn[
        \centering
        \Large
        \textbf{\thetitle}\\
        \vspace{1.4em}Supplementary Material \\
        \vspace{1.0em}
       ] 
   }

\author{Mor Shpigel Nacson\thanks{Equal contribution. Corresponding author: mor.shpigel@gmail.com.} \thanks{Work done during an Amazon internship.} \quad Aviad Aberdam\footnotemark[1] \quad Roy Ganz \quad Elad Ben Avraham\\ Alona Golts \quad Yair Kittenplon \quad Shai Mazor \quad Ron Litman\\
AWS AI Labs%
}

\begin{document}
\maketitle
\begin{abstract}
    
Vision-Language Models (VLMs) excel in diverse visual tasks but face challenges in document understanding, which requires fine-grained text processing. While typical visual tasks perform well with low-resolution inputs, reading-intensive applications demand high-resolution, resulting in significant computational overhead. Using OCR-extracted text in VLM prompts partially addresses this issue but underperforms compared to full-resolution counterpart, as it lacks the complete visual context needed for optimal performance.
We introduce \emph{DocVLM}, a method that integrates an OCR-based modality into VLMs to enhance document processing while preserving original weights. Our approach employs an OCR encoder to capture textual content and layout, compressing these into a compact set of learned queries incorporated into the VLM. Comprehensive evaluations across leading VLMs show that DocVLM significantly reduces reliance on high-resolution images for document understanding.
In limited-token regimes (448$\times$448), DocVLM with 64 learned queries improves DocVQA results from 56.0\% to 86.6\% when integrated with InternVL2 and from 84.4\% to 91.2\% with Qwen2-VL. In LLaVA-OneVision, DocVLM achieves improved results while using 80\% less image tokens. The reduced token usage allows processing multiple pages effectively, showing impressive zero-shot results on DUDE and state-of-the-art performance on MP-DocVQA, highlighting DocVLM’s potential for applications requiring high-performance and efficiency.

\end{abstract}    
\section{Introduction}
\label{sec:intro}



\begin{figure}[ht] 
  \centering
  \includegraphics[width=0.98\linewidth]{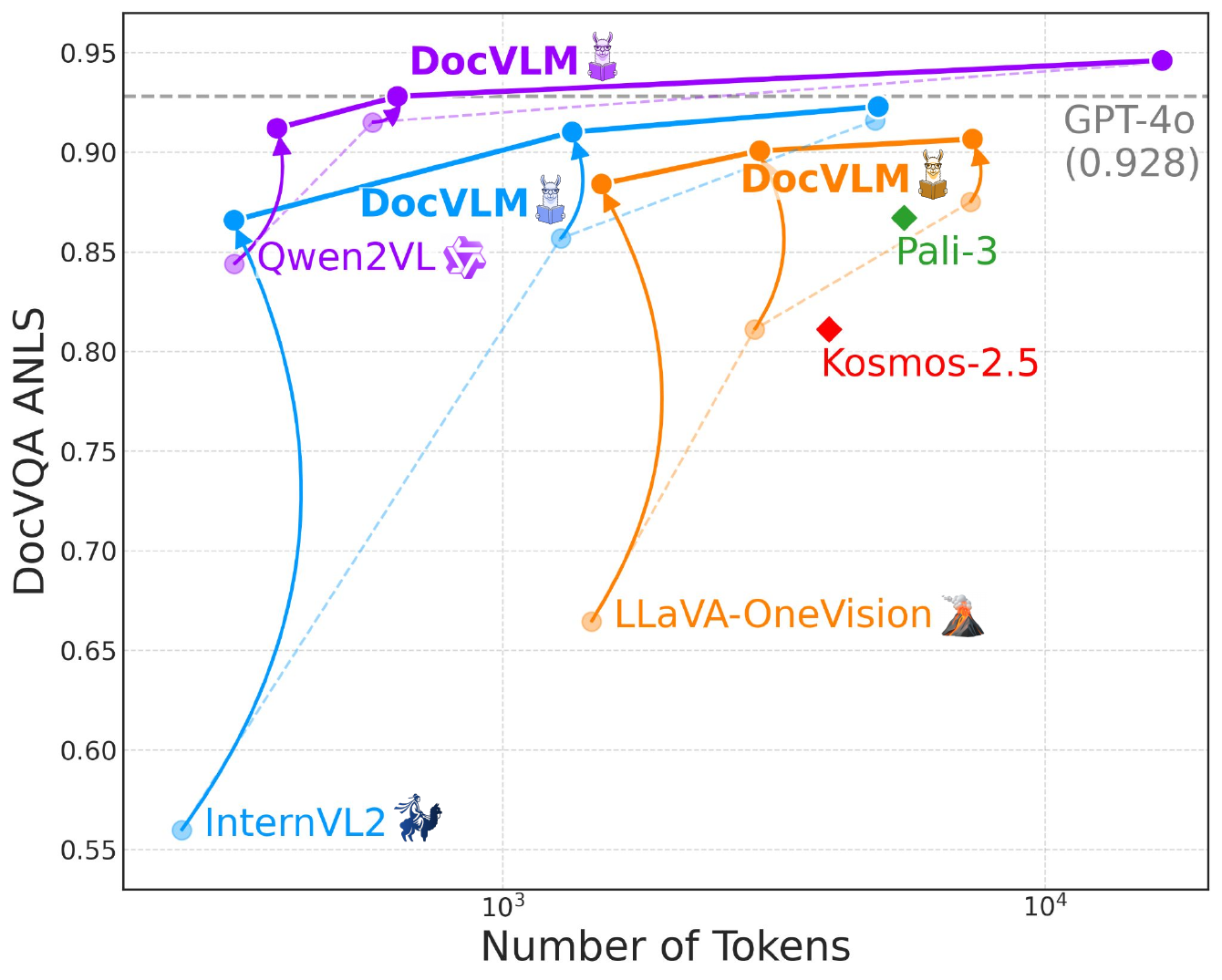}
  \caption{
  \textbf{DocVLM enhances VLMs' reading capabilities.}
  Integrating DocVLM (solid lines) in top-performing VLMs (dashed lines) consistently improves the performance across all token budgets, frequently surpassing the baseline at higher token counts.
  }
    \label{fig:teaser}
  \vspace{-0.4cm}
\end{figure}


The ability to read and interpret text within images is crucial for numerous real-world applications, particularly in document understanding. This field encompasses diverse document types, from dense-text to infographics and multipage documents~\cite{mathew2021docvqa,jaume2019,mathew2022infographicvqa,masry2022chartqa,tito2023hierarchical}, involving tasks that require capabilities in text comprehension, layout understanding, and visual interpretation. Despite progress in VLMs, processing such documents remains challenging \cite{liu2024ocrbenchhiddenmysteryocr}, primarily due to the tension between resolution requirements and computational efficiency. While typical computer vision tasks achieve good performance with low-resolution inputs (typically 224×224 or 336×336 pixels), document analysis demands significantly higher resolutions, resulting in substantial computational overhead~\cite{zhang2024internlm,beyer2024paligemma}.

To address these challenges, some methods incorporate OCR-extracted text directly into language model prompts \cite{chen2023pali, baechler2024screenai}. However, this approach typically lags behind OCR-free methods with full-resolution, as it fails to capture crucial visual context and layout information~\cite{wang2023layout}. Moreover, it often produces long sequence inputs, increasing latency and computational costs, especially for dense documents.

Alternatively, recent VLMs~\cite{li2024llava,dong2024internlm,wang2024qwen2} have introduced specialized mechanisms to reduce visual token count, such as image resizing, tiling limits, and feature downsampling.
However, when applied to document understanding, these methods result in significant performance degradation, creating an undesirable trade-off between computational efficiency and accuracy, as demonstrated in \cref{fig:teaser}.
These limitations underscore the need for a more efficient approach that maintains high-performance document understanding while reducing computational demands.

To overcome these limitations, we introduce DocVLM, a model-agnostic method that enhances VLMs' reading ability by utilizing OCR information effectively. Our approach employs an OCR encoder to capture both contextual and layout details from OCR-extracted text, compressing these encodings into 64 learned queries. These queries are then projected and fed directly into the LLM part of the VLM alongside visual features. 
Unlike some previous methods~\cite{alayrac2022flamingo,laurenccon2024matters,li2023blip,Dai2023InstructBLIPTG}, our compression mechanism avoids the need for a separate compression module or alterations to the LLM architecture.

We demonstrate DocVLM's effectiveness across multiple state-of-the-art VLMs: LlaVA-OneVision \cite{li2024llava}, InternVL2 \cite{chen2024internvl2}, and Qwen2-VL \cite{wang2024qwen2}, each employing a unique image token reduction technique. As illustrated in Figure~\ref{fig:teaser}, our method significantly improves performance, particularly in low input token regimes. Our experiments demonstrate consistently, across all studied VLMs and visual token budgets, that DocVLM's OCR encoder not only outperforms the baseline of inserting OCR words into VLMs but also achieves this superior performance when compressed to just 64 tokens. This dual advantage of improved performance and reduced token usage allows for better utilization of fixed token budgets, enabling the allocation of more tokens for visual processing and further enhancing overall performance.

Importantly, this reduction in sequence length improves the scalability of our approach, enabling its application to multipage document understanding tasks without additional training. We demonstrate that DocVLM can be seamlessly extended to long-context scenarios, such as multipage DocVQA \cite{tito2023hierarchical,van2023document}. While current OCR-free approaches struggle with the overwhelming amount of data in multipage documents \cite{hu2024mplug}, our method achieves strong zero-shot performance on DUDE and surpasses the current state-of-the-art results on MP-DocVQA (86.3\% vs. 80.3\%), despite not being trained on multipage data.

\noindent\textbf{Main contributions:} 
\begin{itemize} 
\item DocVLM, a model-agnostic method that efficiently integrates OCR information into VLMs, capturing both text and layout without complex integration techniques.
\item A compression mechanism that reduces OCR data into a compact set of typically 64 learned queries, significantly reducing computational overhead.
\item Demonstration of DocVLM's effectiveness across different VLM architectures, LlaVA-OneVision, InternVL2, and Qwen2-VL, showing significant performance improvements in low input token regimes (448$\times$448).
\item Extension of DocVLM to long-context tasks, achieving strong zero-shot performance on DUDE and SOTA results on MP-DocVQA without multipage training data.
\end{itemize}

\section{Related Work}

\noindent\textbf{OCR-free Document VLMs.} Early VLMs\cite{alayrac2022flamingo, ganz2023towards, li2023blip, liu2024visual, peng2023kosmos, zhu2023minigpt, ye2023mplug, bai2023qwen, Ganz_2024_CVPR} used relatively small image sizes (e.g., $224 \times 224$ and $336 \times 336$), performing well on natural-image tasks but falling short in document understanding. To address this, recent approaches enhance document understanding capabilities by operating on high-resolution images, developing various strategies to manage the resulting computational burden. Direct processing methods like Donut~\cite{kim2021donut}, PaLI-X~\cite{chen2023palix}, and Qwen2-VL~\cite{wang2024qwen2} attempt full-resolution processing but often resort to image resizing for computational feasibility. Tile-based approaches such as UReader~\cite{ye2023ureader} and InternVL2~\cite{chen2024internvl2} improve efficiency by processing image tiles independently. Other methods, exemplified by LLaVA-1.5~\cite{li2024llava} and LLaVA-OneVision~\cite{liu2024llavanext}, process full-scale images as tiles but downsample the resulting visual features. While these approaches offer different trade-offs, our experiments show their performance deteriorates significantly when constrained to fewer visual tokens.

\noindent\textbf{OCR-Enhanced Document Understanding.} The widespread availability of efficient, open-source OCR models and cost-effective commercial solutions has driven broad adoption of OCR-based approaches in document understanding~\cite{litman2020scatter,nuriel2022textadain,aberdam2021sequence,aberdam2022multimodal,aberdam2023clipter,abramovich2025visfocus,kil2023towards, ye2023deepsolo, ye2022dptext, fujitake2024dtrocr, ronen2022glass}. Several recent works~\cite{Dai2023InstructBLIPTG,bai2023qwen, chen2023pali, chen2023palix} have explored integrating OCR systems with VLMs by feeding extracted text directly into the language model component. Some approaches further enhance this integration by incorporating spatial layout information~\cite{wang2023docllm,wang2023layout,biten2022latr,ganz2023towards,blau2024gram}. While these methods reduce the computational burden of processing high-resolution images, they currently lag behind OCR-free approaches in performance. Additionally, they face challenges with lengthy input sequences, particularly in multipage settings, which can increase latency and computational costs.

\noindent\textbf{Document Representation Compression.} To address efficiency challenges in processing documents, various compression techniques have been developed. 
For OCR-enhanced approaches, \cite{blau2024gram} proposed compressing the OCR signal in multi-page documents using a Compression Transformer, which, despite improving performance in multi-page benchmarks, introduces significant complexity to the system.
In the OCR-free setting, generic VLM approaches like Q-former~\cite{li2023blip} and Resampler~\cite{alayrac2022flamingo} compress visual features but struggle with text-dense images. Document-specific methods such as TokenPacker~\cite{li2024tokenpacker} and DocCompressor~\cite{hu2024mplug} achieve effective visual compression but show reduced performance on document understanding tasks. 
In contrast, rather than compressing high-resolution visual inputs, our DocVLM method operates on lower-resolution images and compresses the extensive OCR signal, including textual and layout information, into a compact set of features (typically 64).

\section{Our Method} \label{sec: method} 

We present \emph{DocVLM}, a model-agnostic approach that enhances VLMs' reading capabilities, enabling operation with lower-resolution inputs while maintaining or improving document understanding accuracy. Our design preserves the base VLM weights, facilitating easy integration across different model architectures and providing flexibility to balance OCR and visual tokens during inference.

\subsection{Architecture} \label{subsec:architecture}

\begin{figure}[t] 
  \centering
  \includegraphics[width=0.99\linewidth]{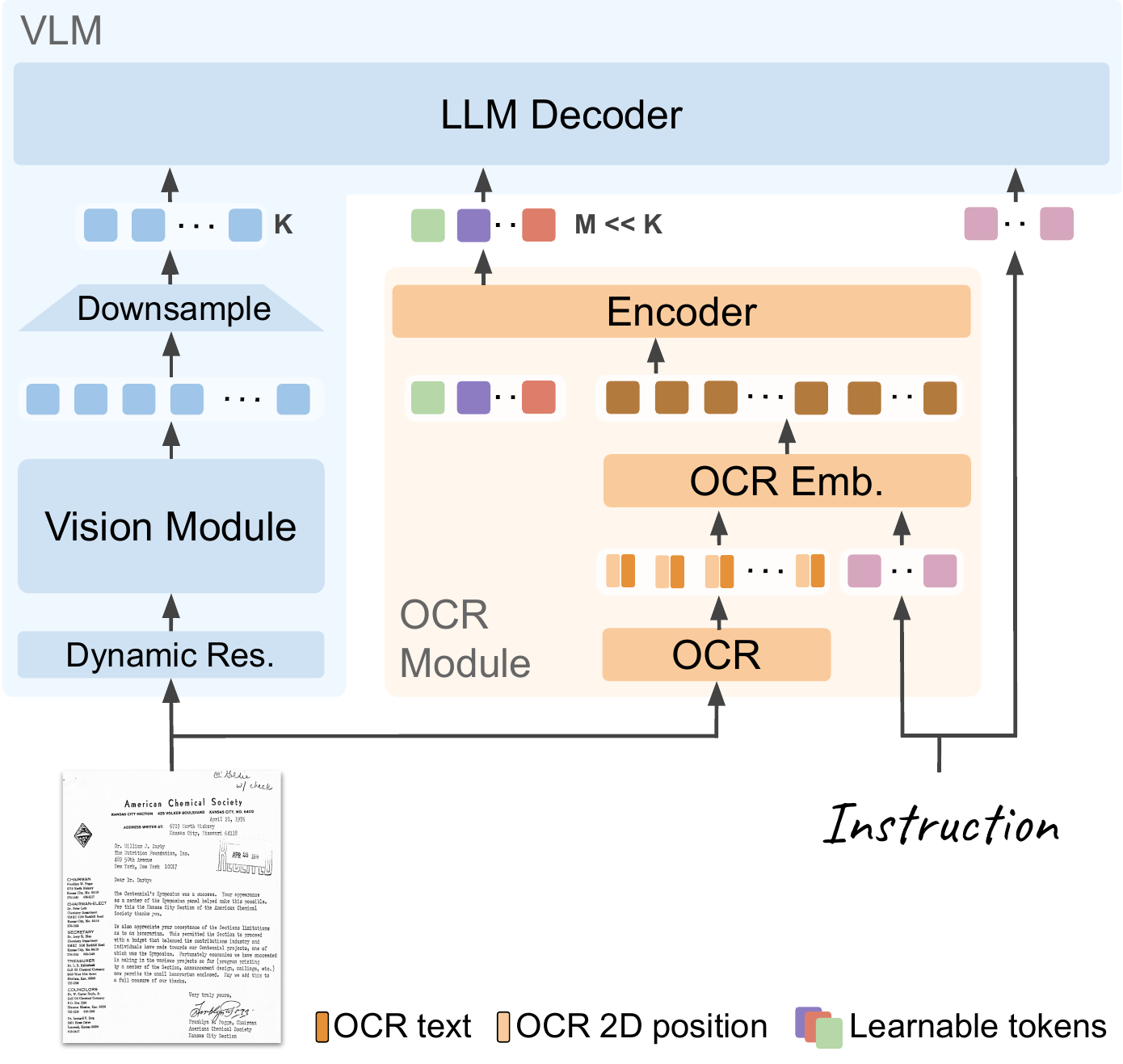}
  \caption{
  \textbf{DocVLM Architecture}. 
    DocVLM enhances document understanding in frozen VLMs by integrating an OCR module with a query compression mechanism.
    By condensing OCR data into $\text{M}=64$ learnable tokens, DocVLM effectively complements visual information, surpassing the VLM’s inherent approaches of increasing image resolution or visual feature dimensions.
    \label{fig:method}
  }
    \vspace{-0.3cm}
\end{figure}

Our method introduces two main components that complement existing VLM architectures: an OCR encoder that processes OCR extracted text and layout information, and a query compression mechanism that distills this information into a compact representation. We integrate these components with pre-trained VLMs, which employ various strategies to control the number of visual tokens for efficient processing. Figure \ref{fig:method} illustrates the overall architecture.

\vspace{-0.3cm}
\paragraph{OCR Encoder Architecture}
We utilize DocFormerV2~\cite{appalaraju2024docformerv2}, a T5-based encoder-decoder~\cite{raffel2020exploring} designed for document understanding, which incorporates vision, language, and spatial features. Specifically, we leverage only the encoder component, which comprises 344 million parameters, and omit its visual branch to eliminate redundancy with the VLM's vision capabilities and reduce computational complexity. The encoder processes two types of inputs: user instructions and OCR data from an OCR system, which consists of textual tokens and their corresponding 2D positional information~\cite{appalaraju2021docformer,appalaraju2024docformerv2,biten2022latr,hsu2024m3t,ganz2023towards,blau2024gram}.

\vspace{-0.1cm}
\paragraph{Query Compression Mechanism} 
To efficiently integrate OCR information into VLMs, we introduce an instruction-aware compression mechanism that distills the OCR encoder's output into a compact set of learned queries. This mechanism significantly reduces the input sequence length for the language model while preserving essential document information.
The compression process utilizes $M$ learnable queries $\mathbf{Q}$ (typically $M=64$), initialized randomly following the OCR encoder embeddings' distribution.
These queries are processed by the OCR encoder alongside two types of embeddings: OCR embeddings ($\mathbf{E}_{\text{OCR}}$),  which encode both OCR tokens and their bounding boxes, and instruction embeddings ($\mathbf{E}_{\text{Instructions}}$). The encoding process can be represented as:
\begin{equation*}
    \operatorname{Encoder}([\mathbf{E}_{\text{OCR}}, \mathbf{E}_{\text{Instructions}}, \mathbf{Q}]).
\end{equation*}
From the encoder output, we retain only the $M$ features corresponding to the learned queries. These compressed features are then projected to match the VLM's hidden dimension and concatenated with the visual tokens before entering the language model. This compression significantly reduces the LLM's input sequence length, enabling either more efficient processing or, under a fixed token budget, allocation of additional tokens to visual features.

\vspace{-0.3cm}
\paragraph{Vision Process}
OCR-free VLMs employ different visual processing methods and various strategies to control the number of visual tokens, aiming to reduce the computational cost of processing high-resolution images needed for document understanding. These approaches can be grouped into three main paradigms:
\begin{enumerate}
    \item 
    \textbf{Full Image Processing with Image Resizing (e.g., Qwen2-VL~\cite{wang2024qwen2}):} 
    In this approach, the model processes the entire image as a single input, controlling the number of visual tokens by resizing the image to a fixed or range-constrained image resolutions.
    While this image processing preserves global context, it incurs quadratic computational complexity with respect to the number of visual tokens.
    
    \item \textbf{Patch-Based Processing with Controlled Tile Count (e.g., InternVL2~\cite{chen2024internvl2}):} 
    This strategy segments the input image into spatial tiles, processing each independently, and controls the number of visual tokens by limiting the number of tiles.
    While most implementations incorporate a low-resolution global view, the primary focus on local processing may compromise global context understanding.
    The computational complexity for this approach scales linearly with respect to the number of tiles as the image increase. The result is improved memory efficiency compared to processing the full image at once, especially for large images.
    
    \item \textbf{Full-Scale Processing with Feature Downsampling (e.g., LlaVA-OneVision~\cite{liu2024llavanext}):}
    Some VLMs initially process full-scale images, but then downsample the visual features into maximal token count before feeding them to the LLM.
    While this method captures both global and local context, it introduces significant computational overhead during initial full-scale processing.
\end{enumerate}
\vspace{0.1em}
\noindent Experimental results confirm that our OCR query compression mechanism significantly enhances document understanding capabilities across all three visual processing strategies, demonstrating its effectiveness as a universal enhancement to existing VLM architectures.

\subsection{Training Strategy} 
\label{sec:training}

Our training strategy aims to integrate the OCR modality into existing VLMs while preserving their core strengths. To achieve this, we keep the VLM completely frozen throughout the process and train only the newly introduced OCR components, i.e., the learnable queries, the OCR encoder, and the projection layer. We employ a two-stage training strategy to gradually incorporate the OCR modality into the pre-trained VLM:

\noindent\textbf{Stage I: OCR-LLM Alignment.} During this stage, we withhold image input from the VLM, forcing the model to rely solely on the newly introduced OCR modality. This approach ensures full utilization of OCR data, aligns OCR components with the LLM input space, and reduces sequence length, improving training efficiency. Given the focus on text input, our dataset selection concentrates on text-related tasks. We begin by training only the randomly initialized components: the learnable queries and projection layer. This allows these components to adapt without disrupting the pretrained OCR encoder. Subsequently, we unfreeze the OCR encoder for fine-tuning, enabling a more comprehensive alignment of the entire encoder to the VLM.

\noindent\textbf{Stage II: Vision Alignment.} In this final stage, we incorporate visual information extracted from the visual encoder, encouraging the OCR components to complement the visual features. Our experiments reveal this stage has a particularly strong effect when using fewer learned queries, allowing the compressed OCR information to better supplement the information acquired from the visual modality (see \cref{sec:ablations}). During this stage, we add more visually focused datasets to the training process. Note that despite our method preserving the original VLM weights, it might implicitly inject bias through prompt tuning. To avoid this, the training data should represent all tasks of interest.

\subsection{Multipage Document Extension} \label{sec:mp}

We conduct our training procedure on single-page data only. However, our approach can be extended to operate on multipage documents. Given a multipage input and its OCR information, the VLM independently processes each page image and concatenates the resulting visual features. For the OCR information, we explore two strategies: \emph{Global Encoding}, which compresses the entire document's OCR information into 64 learnable queries, and \emph{Page-wise Encoding}, which compresses each page's OCR information separately into 64 learnable queries and then concatenates them, resulting in 64 $\times$ number of pages learned queries. After processing the OCR information using either strategy, we feed the resulting compressed OCR features along with the concatenated visual features into the LLM.

Our experiments demonstrate that both approaches are highly effective and efficient in processing multipage documents. Using either approach with a restricted number of visual tokens, we obtain strong zero-shot results on DUDE~\cite{van2023document} and state-of-the-art results on MP-DocVQA \cite{tito2023hierarchical}. The page-wise encoding strategy yields slightly better results for lower numbers of visual tokens.

\section{Experiments}

\begin{table*}[ht]
\centering
\resizebox{\linewidth}{!}{
\begin{tabular}{l c c c c c c c c c}
\toprule
\textbf{Method} & \textbf{\# Tok.} & \textbf{\#P} & {DocVQA} & {TextVQA} & {ST-VQA} & {InfoVQA} & {TextCAPS} & {MP-DocVQA} & {DUDE$^{\star}$} \\
\midrule
\multicolumn{10}{c}{\textbf{No Token Limitations}} \\
\midrule
\rowcolor{gray!25} GPT-4o &  &  & 92.8 & 77.4 & - & 79.2 & - & - & - \\
\rowcolor{gray!25} Gemini 1.5 Pro &  &  & 93.1 & 78.7 & - & 81.0 & - & - & - \\
\rowcolor{gray!25} GPT-4V &  &  & 87.2 & 78.0 & - & 75.1 & - & - & - \\
\rowcolor{gray!15} KOSMOS-2.5-CHAT & 4K & 1.3B & 81.1 & 40.7 & - & 41.3 & - & - & - \\
\rowcolor{gray!15} TextSquare & 2.5K & 8.6B & 84.3 & 66.8 & - & 51.5 & - & - & - \\
\rowcolor{gray!15} ScreenAI & 3.5K & 5B & 87.8 & - & - & 57.8 & - & 72.9 & - \\
\rowcolor{gray!15} ScreenAI+OCR & 4.3K & 5B & 89.9 & - & - & 65.9 & - & 77.1 & - \\
\rowcolor{gray!15} Pali-3 & 5.5K & 5B & 86.7 & 79.5 & 84.1 & 57.8 & 158.8 & - & - \\
\rowcolor{gray!15} Pali-3+OCR & 6.3K & 5B & 88.6 & 80.8 & 85.7 & 62.4 & 164.3 & - & - \\
\midrule
\multicolumn{10}{c}{\textbf{\# Tokens $\leq$ 1.5k}} \\
\midrule
UReader & 841 & 7B & 65.4 & 57.6 & - & 42.2 & 118.4 & - & - \\
Monkey & 1.3K & 9B & 66.5 & 64.3 & - & 36.1 & 93.2 & - & - \\
TextMonkey & 768 & 9B & 73.0 & 65.9 & - & 28.6 & - & - & - \\
Vary & 256 & 7B & 76.3 & - & - & - & - & - & - \\
DocOwl2 & 324 & 8B & 80.7 & 66.7 & - & 46.4 & 131.8 & {69.4} & \textcolor{gray}{46.8} \\
GRAM & 900 & 1B & 85.3 & - & - & - & - & {80.3} & \textcolor{gray}{51.2} \\
GRAM$_{\text{C-Former}}$ & 256 & 1B & 87.6 & - & - & - & - & {77.6} & \textcolor{gray}{45.5} \\
DocFormer v2 & 1K & 1B & 87.8 & 64.0 & 71.8 & 48.8 & - & {76.4} & \textcolor{gray}{48.4} \\
\hdashline
\rowcolor{gray!15} LLaVA-OneVision & 7K & 7B & 87.5 & 76.1 & 71.1 & 68.8 & 138.0 & OOM & OOM \\
LLaVA-OneVision & 1.5K & 7B & 66.5 & 72.1 & 70.6 & 45.6 & 112.9 & 41.8 & 28.7 \\
\rowcolor{cyan!10} \textbf{DocVLM$_{\text{LLaVA-OneVision}}$ (Ours)} & 1.5K & 7B & 88.4 & 76.9 & 70.8 & 61.0 & 145.3 & 77.9 & 43.8 \\
\hdashline
\rowcolor{gray!15} InternVL 2 & 3.1K & 8B & 91.6 & 77.4 & - & 74.8 & - & OOM & OOM \\
InternVL 2 & 256 & 8B & 56.0 & 65.7 & 65.7 & 38.4 & 51.1 & 51.0 & 30.5 \\
\rowcolor{cyan!10} \textbf{DocVLM$_{\text{InternVL2}}$ (Ours)} & 320 & 8B & 86.6 & 71.2 & 74.3 & 57.6 & 119.4 & 76.2 & 43.3 \\
InternVL 2 & 1280 & 8B & 85.7 & 75.5 & 68.3 & 61.5 & 43.7 & 78.1 & 42.2 \\
\rowcolor{cyan!10} \textbf{DocVLM$_{\text{InternVL2}}$ (Ours)} & 1344 & 8B & 91.0 & 76.7 & 76.7 & 65.4 & 123.4 & 81.8 & 45.6 \\
\hdashline
\rowcolor{gray!15} Qwen2-VL & 16k & 7B & 94.5 & 84.3 & 70.7 & 76.5 & 150.2 & OOM & OOM \\
Qwen2-VL & 320 & 7B & 84.4 & 78.0 & 70.1 & 54.1 & 142.1 & 73.0 & 41.5 \\
\rowcolor{cyan!10} \textbf{DocVLM$_{\text{Qwen2-VL}}$ (Ours)} & 320 & 7B & 91.2 & 79.6 & 76.5 & 61.2 & 144.3 & 81.7 & 46.1 \\
Qwen2-VL & 576 & 7B & 91.5 & 82.3 & 70.5 & 65.3 & 145.0 & 82.1 & 45.9 \\
\rowcolor{cyan!10} \textbf{DocVLM$_{\text{Qwen2-VL}}$ (Ours)} & 576 & 7B & \textbf{92.8} & \textbf{82.8} & \textbf{79.8} & \textbf{66.8} & \textbf{150.4} & \textbf{84.5} & \textbf{47.4} \\
\bottomrule
\end{tabular}}
\caption{
\textbf{Comparison with State-of-the-Art Methods.} Performance evaluation of DocVLM against state-of-the-art approaches on document understanding benchmarks. Results are categorized into unconstrained models and those with a 1.5k token limit. In the constrained token regime, DocVLM consistently enhances the performance of baseline VLMs across various tasks and visual token budgets. Notably, DocVLM paired with Qwen2-VL (576 tokens) achieves superior performance across all evaluated datasets, including state-of-the-art zero-shot accuracy on DUDE.
'$\star$' indicates zero-shot evaluation, with grey entries denoting non-zero-shot results.
\label{tab:SOTA table}
}
\vspace{-0.3cm}
\end{table*}

\subsection{Experimental Setting}
\label{subsec:experimental_setting}

\noindent\textbf{Model Integration:}
We evaluate DocVLM through integration with three leading open-source VLMs: LLaVA-OneVision~\cite{li2024llava}, InternVL2~\cite{chen2024internvl2}, and Qwen2-VL~\cite{wang2024qwen2}. As discussed in \cref{subsec:architecture}, these models employ distinct token reduction strategies, enabling us to assess DocVLM's effectiveness across different visual processing approaches. 

\noindent\textbf{Training:}
Our training protocol, detailed in \cref{sec:training}, employs a two-phase strategy. The initial phase focuses on text-centric tasks using datasets spanning document understanding (DocVQA~\cite{mathew2021docvqa}, InfoVQA~\cite{mathew2022infographicvqa}), scene text analysis (ST-VQA~\cite{biten2019scene}, TextVQA~\cite{singh2019towards}, OCR-VQA~\cite{mishra2019ocr}), and specialized tasks (ChartQA~\cite{masry2022chartqa}, TextCaps~\cite{sidorov2020textcaps}, TAT-DQA~\cite{zhu2022towards}). The subsequent vision alignment phase incorporates additional visual-centric datasets: COCO Caption~\cite{chen2015microsoft} and VQA-V2~\cite{balanced_vqa_v2}.

\noindent\textbf{Evaluation:} For evaluation, we focus on five key benchmarks: DocVQA, TextVQA, ST-VQA, InfoVQA, and TextCaps. Results are reported on test sets where available, with TextVQA and TextCaps evaluated on validation sets due to test server restrictions. We use ANLS as the evaluation metric for all datasets, except TextVQA, which uses VQAScore, and TextCaps, which uses CIDEr. To demonstrate DocVLM's generalization capabilities, we conduct zero-shot evaluation on multipage document understanding benchmarks: DUDE~\cite{van2023document} and MP-DocVQA~\cite{tito2023hierarchical}. This zero-shot performance is particularly noteworthy as our model is trained exclusively on single-page documents. Additional implementation details, including hyperparameters and optimization strategies, are provided in the supplementary.

\subsection{State-of-the-art Comparisons}

\begin{figure*}[t] 
  \centering
  \includegraphics[width=0.99\linewidth]{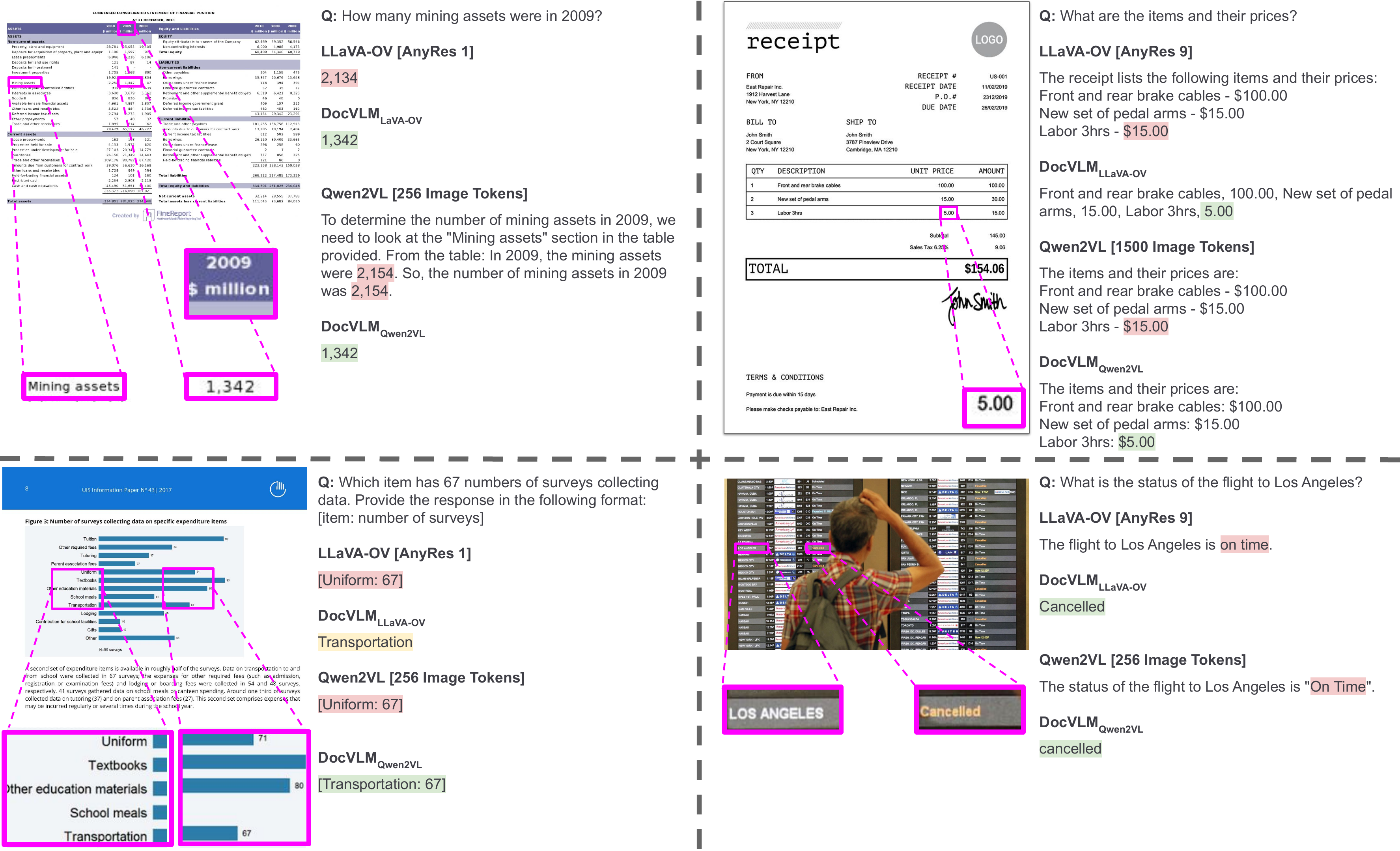}
  \caption{
    \textbf{Qualitative Results.} Representative examples of DocVLM's performance across diverse document formats, from dense text to infographics and scene text.
    Our model successfully handles complex layouts, dense content, and presents instruction-following capabilities without explicit training on such datasets.
    Each example includes an image-instruction pair with baseline and DocVLM predictions.
    }
    \label{fig:qualitative results}
  \vspace{-0.3cm}
\end{figure*}

Table \ref{tab:SOTA table} presents comprehensive comparisons between DocVLM and other state-of-the-art methods across various document understanding benchmarks, highlighting DocVLM’s ability to improve performance under token constraints.
We categorize the results into two main groups: methods without token constraints (both closed and open-source models) and those operating under a 1.5k token limit. We mainly focused on models with around 7B parameters and include methods using an OCR system such as Pali-3~\cite{beyer2024paligemma} ScreenAI~\cite{baechler2024screenai}, DocFormerV2~\cite{appalaraju2024docformerv2}, and GRAM~\cite{blau2024gram}.

To evaluate DocVLM's effectiveness under token constraints, we integrated it with three baseline models: LlaVA-OneVision \cite{li2024llava}, InternVL2 \cite{chen2024internvl2}, and Qwen2-VL \cite{wang2024qwen2}, each configured to operate within the 1.5k token limit. For LlaVA-OneVision, we utilized the minimal visual token configuration (single visual features tile). InternVL2 was tested with both single-tile (256 tokens) and four-tile (1280 tokens) configurations, while Qwen2-VL was evaluated with 256 and 512 visual tokens, corresponding to image sizes of $448 \times 448$ and $616 \times 616$ respectively.

Under the 1.5k token constraint, which is essential for real-world applications, incorporating DocVLM with each of these baseline models yields substantial and consistent improvements. Notably, these improvements persist even in looser token regimes, such as InternVL2 with 1280 visual tokens and Qwen2-VL with 576 visual tokens. Within this constraint, our Qwen2-VL variant with DocVLM, using just 576 tokens, achieves state-of-the-art performance across all benchmarks: 92.8\% on DocVQA, 82.8\% on TextVQA, 79.8\% on ST-VQA, 66.8\% on InfoVQA, and a CIDEr score of 150.4 on TextCAPS. 

DocVLM also demonstrates exceptional capability in handling multipage documents. Using the same 576-token configuration described earlier, it achieves 84.5\% accuracy on MP-DocVQA, surpassing previous state-of-the-art results. Furthermore, this setup shows robust generalization capabilities with a zero-shot performance of 47.4\% accuracy on the DUDE dataset, despite not being specifically trained for multipage document processing.

\subsection{Qualitative Results}

Figure \ref{fig:qualitative results} illustrates DocVLM's enhanced capabilities through representative examples that demonstrate three key strengths of our method: (1) improved reading comprehension in complex document layouts, (2) effective handling of dense textual content despite using compressed representations, and (3) preserved and enhanced instruction-following capabilities.

The examples span diverse document types, from dense text documents to infographics and scene text, showcasing DocVLM's versatility. In the infographic example, DocVLM not only preserves but enhances the base model's instruction-following capabilities, despite our OCR component not being explicitly trained on instruction-following datasets like \cite{tanaka2024instructdoc}. This demonstrates that our compression mechanism successfully retains crucial textual and layout information while significantly reducing token usage.

\subsection{Scaling to Multipage Documents}

Building on the promising multipage results in \cref{tab:SOTA table}, we conduct an in-depth analysis of different DocVLM configurations for multipage document understanding. This analysis focuses on the Qwen2-VL base model, tested on the MP-DocVQA dataset with documents up to 20 pages long -- a scale that most other state-of-the-art methods struggle to handle due to token limitations.

Table \ref{tab: multipage results} compares four OCR integration strategies used during inference in multipage scenarios:
\begin{itemize}
    \item Baseline: vision-only input (no additional tokens)
    \item Direct OCR word insertion: up to 800 tokens per page
    \item Global OCR encoding: 64 tokens total
    \item Page-wise OCR encoding: 64 tokens per page 
\end{itemize}

Results indicate consistent improvements over the baseline across all image resolutions (256, 512, and 1024 tokens), with only minimal additional token usage -- just 64 tokens for the entire document in the global encoding case.
Notably, at 256 visual tokens per page, both page-wise encoding (82.4\%) and global encoding (81.7\%) outperform direct OCR word insertion (79.1\%) while using significantly fewer tokens.

Our best configuration achieves state-of-the-art performance of 86.3\% ANLS, significantly outperforming specialized multipage models like GRAM (80.3\%). This is particularly impressive considering that DocVLM was trained exclusively on single-page inputs, demonstrating strong zero-shot generalization to multipage scenarios.

The comparison between encoding strategies reveals that page-wise encoding consistently outperforms global encoding at lower visual token counts, providing a +0.7\% improvement for both 256 and 512 image tokens per page. This advantage diminishes at 1024 tokens where both achieve identical performance (86.3\% ANLS). Remarkably, DocVLM with page-wise encoding matches or even outperforms the baseline Qwen2-VL using twice as many visual tokens, highlighting the efficiency of our approach.

\begin{table}[t]
\centering
\resizebox{\linewidth}{!}{
\begin{tabular}{llcccc}
\toprule
\textbf{Method} & \textbf{LLM OCR Input} & \textbf{Image Tok.} & \textbf{OCR Tok.} & \textbf{ANLS} \\
\midrule
DocOwl2 \cite{hu2024mplug} & -- & $324\times \text{pg}$ & -- & 69.4 \\
GRAM$_{\text{C-Former}}$  \cite{blau2024gram} & -- & $100\times \text{pg}$ & 256 & 77.6 \\
GRAM  \cite{blau2024gram} & -- & $100\times \text{pg}$ & $800\times \text{pg}$ & 80.3 \\
\hdashline
\multirow{2}{*}{Qwen2-VL} & -- & $256\times \text{pg}$ & -- & 73.0 \\
 & OCR Words & $256\times \text{pg}$ & $800 \times \text{pg}$ & 79.1 \\
\rowcolor{cyan!10} \textbf{$\text{DocVLM}_{\text{Qwen2-VL}}$} & Global Encoding & $256\times \text{pg}$ & 64 & 81.7 \\
\rowcolor{cyan!10} \textbf{$\text{DocVLM}_{\text{Qwen2-VL}}$} & Page-wise Encoding & $256\times \text{pg}$ & $64\times \text{pg}$ & 82.4 \\
Qwen2-VL & -- & $512\times \text{pg}$ & -- & 82.1 \\
\rowcolor{cyan!10} \textbf{$\text{DocVLM}_{\text{Qwen2-VL}}$} & Global Encoding & $512\times \text{pg}$ & 64 & 84.5 \\
\rowcolor{cyan!10} \textbf{$\text{DocVLM}_{\text{Qwen2-VL}}$} & Page-wise Encoding & $512\times \text{pg}$ & $64\times \text{pg}$ & \underline{85.2} \\
Qwen2-VL & -- & $1024\times \text{pg}$ & -- & \underline{85.2} \\
\rowcolor{cyan!10} \textbf{$\text{DocVLM}_{\text{Qwen2-VL}}$} & Global Encoding & $1024\times \text{pg}$ & 64 & \textbf{86.3} \\
\rowcolor{cyan!10} \textbf{$\text{DocVLM}_{\text{Qwen2-VL}}$} & Page-wise Encoding & $1024\times \text{pg}$ & $64\times \text{pg}$ & \textbf{86.3}\\
\bottomrule
\end{tabular}}
\caption{
\textbf{Extension to Multipage.}
Comparison on MP-DocVQA of approaches for incorporating OCR information in multipage document understanding. Both DocVLM multipage extension strategies: global encoding (64 tokens per document) and page-wise encoding (64 tokens per page), outperform previous state-of-the-art methods, notably, without any explicit multipage training.
\label{tab: multipage results}}
\vspace{-0.3cm}
\end{table}

\section{Ablation Study}
\label{sec:ablations}

\paragraph{Impact of OCR Encoding Strategies} To evaluate the impact of OCR encoding compression, we compare three strategies for integrating OCR information: (1) inserting raw OCR words in the original VLM, (2) using DocVLM uncompressed OCR encodings, and (3) DocVLM compressed OCR encodings with 64 learned queries. We evaluate these approaches on the DocVQA test set using three representative model configurations: LLaVA-OneVision with 1.5K visual tokens, and both InternVL2 and Qwen2-VL with 256 visual tokens each.

Results in \cref{tab: components ablation} demonstrate that DocVLM's OCR encodings significantly outperform raw OCR words across all three models while maintaining the same token count. Notably, our compressed encoding approach, using just 64 tokens instead of 800 OCR tokens, preserves most of these improvements while drastically reducing sequence length.
This efficient compression enables a more favorable allocation of the token budget, allowing models to dedicate more tokens to visual processing without compromising OCR effectiveness.
The results validate that DocVLM's compression strategy successfully balances performance with computational efficiency, a key factor for practical applications.

\begin{figure*}
    \centering
    \includegraphics[width=\linewidth]{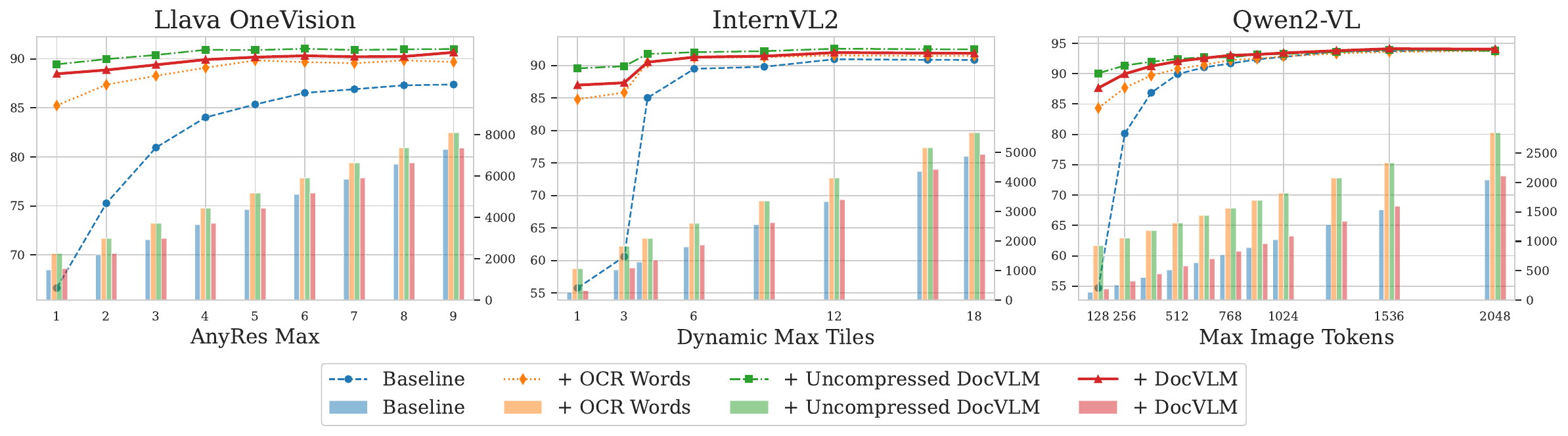}
    \vspace{-0.6cm}
    \caption{
    \textbf{Balancing Performance and Compute.} Analysis of model performance (lines, left y-axis) and token usage (bars, right y-axis) as a function of visual token allocation. Each model employs its inherent token control strategy: AnyRes max for feature downsampling (LLaVA One-Vision), dynamic max tiles (InternVL2), and max image tokens for resolution control (Qwen2-VL).
    The results highlight that DocVLM consistently improves performance with minimal overhead (64 tokens), offering an efficient OCR-visual token allocation.
    }
    \label{fig:ocr fusion gain}
    \vspace{-0.3cm}
\end{figure*}
\begin{table}[t]
\centering
\resizebox{\linewidth}{!}{
\begin{tabular}{l c c c c}
\toprule
\multirow{2}{*}{\textbf{LLM OCR Input}} &  \multirow{2}{*}{\textbf{OCR Tok.}} &  \textbf{LLaVA-OV} & \textbf{InternVL2} & \textbf{Qwen2-VL} \\
 & & 1.5K & 256 & 256 \\
\midrule
OCR Words  & 800 & 85.8 & 84.4 & 89.1\\
\rowcolor{cyan!10} OCR Encoding  & 800 & 89.4 & 89.2 & 91.9 \\
\rowcolor{cyan!10} 64 Compressed Encoding  & 64 & 88.4 & 86.6 & 91.2 \\
\bottomrule
\end{tabular}}
\caption{\textbf{OCR Encoding Strategies.} DocVQA results for inserting OCR information using (1) OCR words (baseline), (2) uncompressed OCR encodings, and (3) 64 compressed OCR encodings.
\label{tab: components ablation}}
\vspace{-0.3cm}
\end{table}

\begin{figure}[t]
    \centering
    \includegraphics[width=0.9\linewidth]{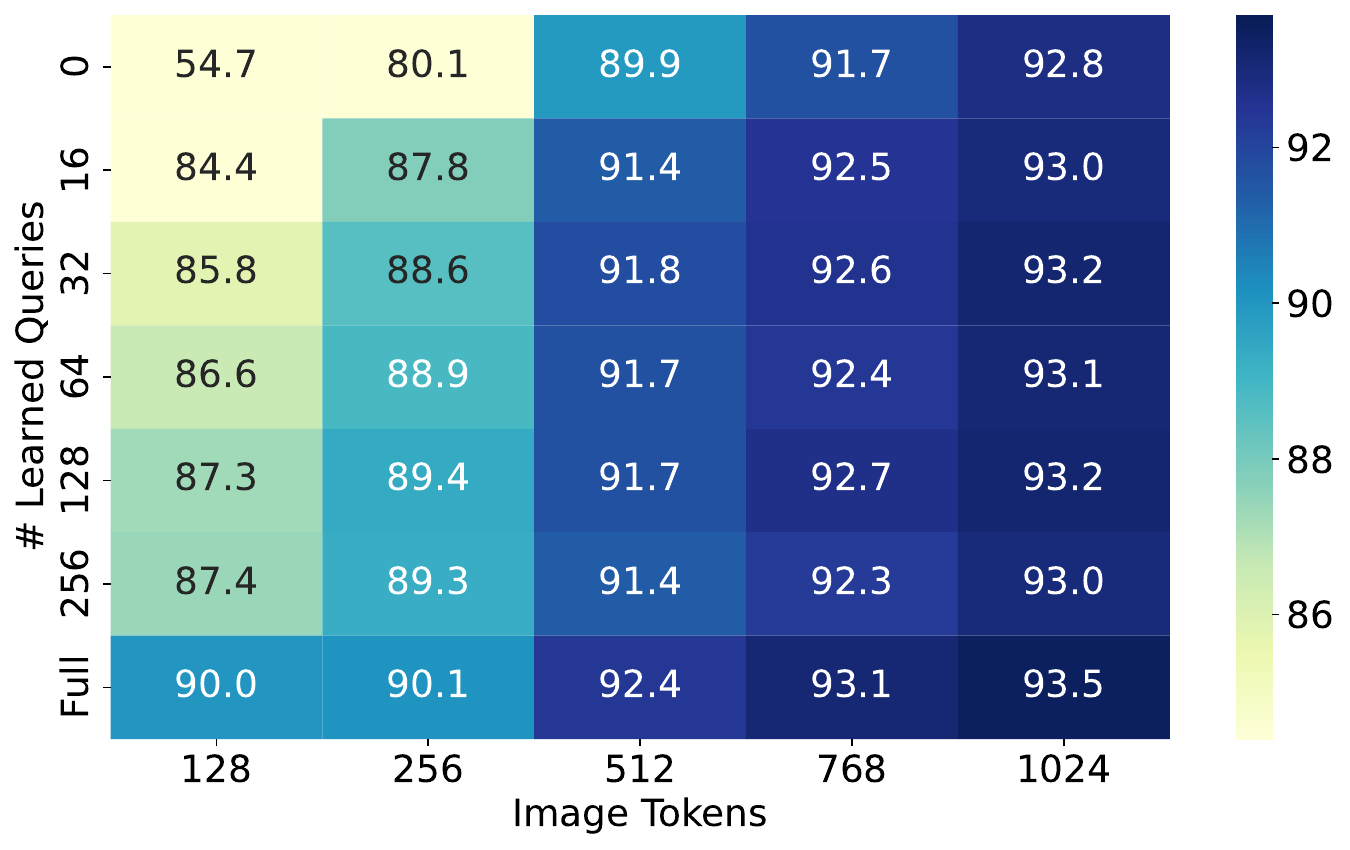}
    \caption{
    \textbf{Compression Levels.} DocVQA validation results for DocVLM integrated with Qwen2-VL across varying OCR and image token budgets. "0" represents the baseline, while "Full" indicates uncompressed encodings.
    }
    \label{fig:anls_vs_tokens_and_lq_heatmap}
    \vspace{-0.3cm}
\end{figure}

\paragraph{Balancing Vision and OCR Token Allocation}
Modern VLMs employ various mechanisms to reduce visual token count, creating an inherent trade-off between computational efficiency and model performance, as discussed in \cref{subsec:architecture}.
We investigate how DocVLM can improve this trade-off by comparing four configurations: (1) baseline VLM without OCR, (2) direct OCR word insertion, (3) DocVLM with uncompressed OCR encodings, and (4) DocVLM with 64 compressed learned queries.

\Cref{fig:ocr fusion gain} presents the performance scores on DocVQA validation (left y-axis) and total token counts (right y-axis) for three VLM architectures, showing how these metrics vary across different visual token allocations. Each model employs a distinct token reduction approach: LLaVA-OneVision controls token count through feature downsampling (AnyRes Max), InternVL2 limits the number of processed image tiles (Dynamic Max Batch), and Qwen2-VL adjusts image resolution to constrain token count. 

Our analysis reveals that integrating OCR information, regardless of the method used, consistently improves performance across all models, with particularly pronounced gains in low visual token regimes. However, uncompressed OCR integration methods, whether through direct word insertion or uncompressed DocVLM encodings, require 800 tokens -- a significant overhead that could otherwise be allocated to visual processing.
For instance, allocating 128 tokens for visual and 800 for OCR in Qwen2-VL achieves 84.3\% using OCR words and 90.1\% using uncompressed encodings. In contrast, using 896 pure visual tokens reaches 92.4\%, demonstrating the potential benefit of allocating more tokens to visual processing.
DocVLM's compression mechanism provides a superior option by requiring only 64 tokens for OCR information while maintaining strong performance. In the above example, our approach using fewer tokens, allocating 768 visual tokens and 64 OCR tokens, reaches 93.0\%, outperforming the 90.1\% obtained with the uncompressed encodings, highlighting DocVLM's effective balance between visual and OCR tokens.

\paragraph{Compression Levels} 
We deepen our analysis of the OCR-visual token trade-off by examining different compression levels in DocVLM integrated with Qwen2-VL. \cref{fig:anls_vs_tokens_and_lq_heatmap} presents ANLS scores across various combinations of visual tokens (128-1024) and learned queries (16-256), including baselines without OCR and with uncompressed encoding. In low visual token regimes, increasing the number of learned queries yields substantial improvements, validating our compression mechanism's effectiveness in capturing relevant OCR information. Notably, even with just 16 learned queries, DocVLM outperforms the baseline across all visual token configurations, offering strong performance with minimal computational overhead.

\paragraph{Training Stages}

\cref{tab: training effect}\, illustrates the impact of the vision alignment stage on ANLS performance for \ourmethod with the LLaVA-OneVision base model on the DocVQA test set, showing results for varying numbers of learned queries (16 to 128) and the uncompressed case. Our two-stage training process initially trains OCR modality components without image input, forcing reliance on OCR data alone, before reintroducing the image in the vision alignment stage to adapt learned queries alongside visual information. The results reveal that vision alignment significantly boosts performance, especially with fewer learned queries: for instance, with 16 learned queries, there's an improvement of +6.2, compared to +0.7 in the uncompressed case. Notably, after vision alignment, \ourmethod with only 16 learned queries outperforms the baseline of OCR words (from Table \ref{tab: components ablation}). These findings underscore the effectiveness of our two-stage training method.


\begin{table}[t]
\centering
\resizebox{\linewidth}{!}{
\begin{tabular}{c c c c c} \toprule \multirow{2}{*}{\textbf{Training Phases}} & \multicolumn{3}{c}{\textbf{Compressed Enc.}} & \textbf{OCR Enc.} \\ & 16 & 64 & 128 & 800 \\ \midrule \rowcolor{cyan!10} {Stage I: OCR-LLM Alignment} & 81.7 & 85.8 & 86.3 & 89.4 \\ \rowcolor{cyan!10} {+ Stage II: Vision Alignment} & \textbf{87.9} & \textbf{88.4} & \textbf{88.4} & \textbf{90.1} \\ \hdashline $\boldsymbol{\Delta}$ & {\color{OliveGreen} \textbf{+6.2}} & {\color{OliveGreen} \textbf{+2.6}} & {\color{OliveGreen} \textbf{+2.1}} & {\color{OliveGreen} \textbf{+0.7}} \\ \bottomrule \end{tabular}
}
\caption{
\textbf{Training Phases.} Second-stage training consistently improves DocVQA performance, both with compressed DocVLM tokens and full OCR encoding.
\label{tab: training effect}}
\vspace{-0.3cm}
\end{table}

\section{Conclusions}

Our results demonstrate that DocVLM can be effectively integrated into various VLMs to enhance their document reading capabilities while significantly reducing their dependency on extensive vision tokens.
The key takeaway is that in token-constrained scenarios, allocating a small portion of tokens to OCR information consistently yields better results than  than using those tokens solely for visual processing.
The effectiveness of our compression mechanism extends beyond single-page documents, as evidenced by achieving state-of-the-art results on MP-DocVQA using the same 64 tokens to represent multiple pages. 
These results establish DocVLM as a practical solution for enhancing document understanding in real-world applications where computational efficiency is crucial.

{
    \small
    \bibliographystyle{ieeenat_fullname}
    \bibliography{biblio}

\begin{thebibliography}{61}
\providecommand{\natexlab}[1]{#1}
\providecommand{\url}[1]{\texttt{#1}}
\expandafter\ifx\csname urlstyle\endcsname\relax
  \providecommand{\doi}[1]{doi: #1}\else
  \providecommand{\doi}{doi: \begingroup \urlstyle{rm}\Url}\fi

\bibitem[Aberdam et~al.(2021)Aberdam, Litman, Tsiper, Anschel, Slossberg, Mazor, Manmatha, and Perona]{aberdam2021sequence}
Aviad Aberdam, Ron Litman, Shahar Tsiper, Oron Anschel, Ron Slossberg, Shai Mazor, R Manmatha, and Pietro Perona.
\newblock Sequence-to-sequence contrastive learning for text recognition.
\newblock In \emph{Proceedings of the IEEE/CVF Conference on Computer Vision and Pattern Recognition}, pages 15302--15312, 2021.

\bibitem[Aberdam et~al.(2022)Aberdam, Ganz, Mazor, and Litman]{aberdam2022multimodal}
Aviad Aberdam, Roy Ganz, Shai Mazor, and Ron Litman.
\newblock Multimodal semi-supervised learning for text recognition.
\newblock \emph{arXiv preprint arXiv:2205.03873}, 2022.

\bibitem[Aberdam et~al.(2023)Aberdam, Bensa{\"\i}d, Golts, Ganz, Nuriel, Tichauer, Mazor, and Litman]{aberdam2023clipter}
Aviad Aberdam, David Bensa{\"\i}d, Alona Golts, Roy Ganz, Oren Nuriel, Royee Tichauer, Shai Mazor, and Ron Litman.
\newblock Clipter: Looking at the bigger picture in scene text recognition.
\newblock In \emph{Proceedings of the IEEE/CVF International Conference on Computer Vision}, pages 21706--21717, 2023.

\bibitem[Abramovich et~al.(2025)Abramovich, Nayman, Fogel, Lavi, Litman, Tsiper, Tichauer, Appalaraju, Mazor, and Manmatha]{abramovich2025visfocus}
Ofir Abramovich, Niv Nayman, Sharon Fogel, Inbal Lavi, Ron Litman, Shahar Tsiper, Royee Tichauer, Srikar Appalaraju, Shai Mazor, and R Manmatha.
\newblock Visfocus: Prompt-guided vision encoders for ocr-free dense document understanding.
\newblock In \emph{European Conference on Computer Vision}, pages 241--259. Springer, 2025.

\bibitem[Alayrac et~al.(2022)Alayrac, Donahue, Luc, Miech, Barr, Hasson, Lenc, Mensch, Millican, Reynolds, et~al.]{alayrac2022flamingo}
Jean-Baptiste Alayrac, Jeff Donahue, Pauline Luc, Antoine Miech, Iain Barr, Yana Hasson, Karel Lenc, Arthur Mensch, Katherine Millican, Malcolm Reynolds, et~al.
\newblock Flamingo: a visual language model for few-shot learning.
\newblock \emph{Advances in neural information processing systems}, 35:\penalty0 23716--23736, 2022.

\bibitem[Appalaraju et~al.(2021)Appalaraju, Jasani, Kota, Xie, and Manmatha]{appalaraju2021docformer}
Srikar Appalaraju, Bhavan Jasani, Bhargava~Urala Kota, Yusheng Xie, and R Manmatha.
\newblock Docformer: End-to-end transformer for document understanding.
\newblock In \emph{Proceedings of the IEEE/CVF international conference on computer vision}, pages 993--1003, 2021.

\bibitem[Appalaraju et~al.(2024)Appalaraju, Tang, Dong, Sankaran, Zhou, and Manmatha]{appalaraju2024docformerv2}
Srikar Appalaraju, Peng Tang, Qi Dong, Nishant Sankaran, Yichu Zhou, and R Manmatha.
\newblock Docformerv2: Local features for document understanding.
\newblock In \emph{Proceedings of the AAAI Conference on Artificial Intelligence}, pages 709--718, 2024.

\bibitem[Baechler et~al.(2024)Baechler, Sunkara, Wang, Zubach, Mansoor, Etter, C{\u{a}}rbune, Lin, Chen, and Sharma]{baechler2024screenai}
Gilles Baechler, Srinivas Sunkara, Maria Wang, Fedir Zubach, Hassan Mansoor, Vincent Etter, Victor C{\u{a}}rbune, Jason Lin, Jindong Chen, and Abhanshu Sharma.
\newblock Screenai: A vision-language model for ui and infographics understanding.
\newblock \emph{arXiv preprint arXiv:2402.04615}, 2024.

\bibitem[Bai et~al.(2023)Bai, Bai, Yang, Wang, Tan, Wang, Lin, Zhou, and Zhou]{bai2023qwen}
Jinze Bai, Shuai Bai, Shusheng Yang, Shijie Wang, Sinan Tan, Peng Wang, Junyang Lin, Chang Zhou, and Jingren Zhou.
\newblock Qwen-vl: A versatile vision-language model for understanding, localization, text reading, and beyond.
\newblock \emph{arXiv preprint arXiv:2308.12966}, 1\penalty0 (2):\penalty0 3, 2023.

\bibitem[Beyer et~al.(2024)Beyer, Steiner, Pinto, Kolesnikov, Wang, Salz, Neumann, Alabdulmohsin, Tschannen, Bugliarello, et~al.]{beyer2024paligemma}
Lucas Beyer, Andreas Steiner, Andr{\'e}~Susano Pinto, Alexander Kolesnikov, Xiao Wang, Daniel Salz, Maxim Neumann, Ibrahim Alabdulmohsin, Michael Tschannen, Emanuele Bugliarello, et~al.
\newblock Paligemma: A versatile 3b vlm for transfer.
\newblock \emph{arXiv preprint arXiv:2407.07726}, 2024.

\bibitem[Biten et~al.(2019)Biten, Tito, Mafla, Gomez, Rusinol, Valveny, Jawahar, and Karatzas]{biten2019scene}
Ali~Furkan Biten, Ruben Tito, Andres Mafla, Lluis Gomez, Mar{\c{c}}al Rusinol, Ernest Valveny, CV Jawahar, and Dimosthenis Karatzas.
\newblock Scene text visual question answering.
\newblock In \emph{Proceedings of the IEEE/CVF international conference on computer vision}, pages 4291--4301, 2019.

\bibitem[Biten et~al.(2022{\natexlab{a}})Biten, Litman, Xie, Appalaraju, and Manmatha]{biten2022latr}
Ali~Furkan Biten, Ron Litman, Yusheng Xie, Srikar Appalaraju, and R Manmatha.
\newblock Latr: Layout-aware transformer for scene-text vqa.
\newblock In \emph{Proceedings of the IEEE/CVF conference on computer vision and pattern recognition}, pages 16548--16558, 2022{\natexlab{a}}.

\bibitem[Biten et~al.(2022{\natexlab{b}})Biten, Tito, Gomez, Valveny, and Karatzas]{biten2022ocr}
Ali~Furkan Biten, Ruben Tito, Lluis Gomez, Ernest Valveny, and Dimosthenis Karatzas.
\newblock Ocr-idl: Ocr annotations for industry document library dataset.
\newblock \emph{arXiv preprint arXiv:2202.12985}, 2022{\natexlab{b}}.

\bibitem[Blau et~al.(2024)Blau, Fogel, Ronen, Golts, Ganz, Ben~Avraham, Aberdam, Tsiper, and Litman]{blau2024gram}
Tsachi Blau, Sharon Fogel, Roi Ronen, Alona Golts, Roy Ganz, Elad Ben~Avraham, Aviad Aberdam, Shahar Tsiper, and Ron Litman.
\newblock Gram: Global reasoning for multi-page vqa.
\newblock In \emph{Proceedings of the IEEE/CVF Conference on Computer Vision and Pattern Recognition}, pages 15598--15607, 2024.

\bibitem[Chen et~al.(2015)Chen, Fang, Lin, Vedantam, Gupta, Doll{\'a}r, and Zitnick]{chen2015microsoft}
Xinlei Chen, Hao Fang, Tsung-Yi Lin, Ramakrishna Vedantam, Saurabh Gupta, Piotr Doll{\'a}r, and C~Lawrence Zitnick.
\newblock Microsoft coco captions: Data collection and evaluation server.
\newblock \emph{arXiv preprint arXiv:1504.00325}, 2015.

\bibitem[Chen et~al.(2023{\natexlab{a}})Chen, Djolonga, Padlewski, Mustafa, Changpinyo, Wu, Ruiz, Goodman, Wang, Tay, et~al.]{chen2023palix}
Xi Chen, Josip Djolonga, Piotr Padlewski, Basil Mustafa, Soravit Changpinyo, Jialin Wu, Carlos~Riquelme Ruiz, Sebastian Goodman, Xiao Wang, Yi Tay, et~al.
\newblock Pali-x: On scaling up a multilingual vision and language model.
\newblock \emph{arXiv preprint arXiv:2305.18565}, 2023{\natexlab{a}}.

\bibitem[Chen et~al.(2023{\natexlab{b}})Chen, Wang, Beyer, Kolesnikov, Wu, Voigtlaender, Mustafa, Goodman, Alabdulmohsin, Padlewski, et~al.]{chen2023pali}
Xi Chen, Xiao Wang, Lucas Beyer, Alexander Kolesnikov, Jialin Wu, Paul Voigtlaender, Basil Mustafa, Sebastian Goodman, Ibrahim Alabdulmohsin, Piotr Padlewski, et~al.
\newblock Pali-3 vision language models: Smaller, faster, stronger.
\newblock \emph{arXiv preprint arXiv:2310.09199}, 2023{\natexlab{b}}.

\bibitem[Chen et~al.(2024)Chen, Wang, Tian, Ye, Gao, Cui, Tong, Hu, Luo, Ma, et~al.]{chen2024internvl2}
Zhe Chen, Weiyun Wang, Hao Tian, Shenglong Ye, Zhangwei Gao, Erfei Cui, Wenwen Tong, Kongzhi Hu, Jiapeng Luo, Zheng Ma, et~al.
\newblock Internvl2: Better than the best—expanding performance boundaries of open-source multimodal models with the progressive scaling strategy, 2024.

\bibitem[Dai et~al.(2023)Dai, Li, Li, Tiong, Zhao, Wang, Li, Fung, and Hoi]{Dai2023InstructBLIPTG}
Wenliang Dai, Junnan Li, Dongxu Li, Anthony Meng~Huat Tiong, Junqi Zhao, Weisheng Wang, Boyang~Albert Li, Pascale Fung, and Steven C.~H. Hoi.
\newblock Instructblip: Towards general-purpose vision-language models with instruction tuning.
\newblock \emph{ArXiv}, abs/2305.06500, 2023.

\bibitem[Dong et~al.(2024)Dong, Zhang, Zang, Cao, Wang, Ouyang, Zhang, Duan, Zhang, Li, et~al.]{dong2024internlm}
Xiaoyi Dong, Pan Zhang, Yuhang Zang, Yuhang Cao, Bin Wang, Linke Ouyang, Songyang Zhang, Haodong Duan, Wenwei Zhang, Yining Li, et~al.
\newblock Internlm-xcomposer2-4khd: A pioneering large vision-language model handling resolutions from 336 pixels to 4k hd.
\newblock \emph{arXiv preprint arXiv:2404.06512}, 2024.

\bibitem[Fujitake(2024)]{fujitake2024dtrocr}
Masato Fujitake.
\newblock Dtrocr: Decoder-only transformer for optical character recognition.
\newblock In \emph{Proceedings of the IEEE/CVF Winter Conference on Applications of Computer Vision}, pages 8025--8035, 2024.

\bibitem[Ganz et~al.(2023)Ganz, Nuriel, Aberdam, Kittenplon, Mazor, and Litman]{ganz2023towards}
Roy Ganz, Oren Nuriel, Aviad Aberdam, Yair Kittenplon, Shai Mazor, and Ron Litman.
\newblock Towards models that can see and read.
\newblock In \emph{Proceedings of the IEEE/CVF international conference on computer vision}, pages 21718--21728, 2023.

\bibitem[Ganz et~al.(2024)Ganz, Kittenplon, Aberdam, Ben~Avraham, Nuriel, Mazor, and Litman]{Ganz_2024_CVPR}
Roy Ganz, Yair Kittenplon, Aviad Aberdam, Elad Ben~Avraham, Oren Nuriel, Shai Mazor, and Ron Litman.
\newblock Question aware vision transformer for multimodal reasoning.
\newblock In \emph{Proceedings of the IEEE/CVF Conference on Computer Vision and Pattern Recognition (CVPR)}, pages 13861--13871, 2024.

\bibitem[Goyal et~al.(2017)Goyal, Khot, Summers{-}Stay, Batra, and Parikh]{balanced_vqa_v2}
Yash Goyal, Tejas Khot, Douglas Summers{-}Stay, Dhruv Batra, and Devi Parikh.
\newblock Making the {V} in {VQA} matter: Elevating the role of image understanding in {V}isual {Q}uestion {A}nswering.
\newblock In \emph{Conference on Computer Vision and Pattern Recognition (CVPR)}, 2017.

\bibitem[Guillaume~Jaume(2019)]{jaume2019}
Jean-Philippe~Thiran Guillaume~Jaume, Hazim Kemal~Ekenel.
\newblock Funsd: A dataset for form understanding in noisy scanned documents.
\newblock In \emph{Accepted to ICDAR-OST}, 2019.

\bibitem[Hsu et~al.(2024)Hsu, Liu, Li, Fujinuma, Nadejde, Niu, Kittenplon, Litman, and Pappagari]{hsu2024m3t}
Benjamin Hsu, Xiaoyu Liu, Huayang Li, Yoshinari Fujinuma, Maria Nadejde, Xing Niu, Yair Kittenplon, Ron Litman, and Raghavendra Pappagari.
\newblock M3t: A new benchmark dataset for multi-modal document-level machine translation.
\newblock \emph{arXiv preprint arXiv:2406.08255}, 2024.

\bibitem[Hu et~al.(2024)Hu, Xu, Zhang, Ye, Yan, Zhang, Jin, Huang, and Zhou]{hu2024mplug}
Anwen Hu, Haiyang Xu, Liang Zhang, Jiabo Ye, Ming Yan, Ji Zhang, Qin Jin, Fei Huang, and Jingren Zhou.
\newblock mplug-docowl2: High-resolution compressing for ocr-free multi-page document understanding.
\newblock \emph{arXiv preprint arXiv:2409.03420}, 2024.

\bibitem[Kil et~al.(2023)Kil, Kim, Seo, Kim, and Kim]{kil2023towards}
Taeho Kil, Seonghyeon Kim, Sukmin Seo, Yoonsik Kim, and Daehee Kim.
\newblock Towards unified scene text spotting based on sequence generation.
\newblock In \emph{Proceedings of the IEEE/CVF Conference on Computer Vision and Pattern Recognition}, pages 15223--15232, 2023.

\bibitem[Kim et~al.(2021)Kim, Hong, Yim, Park, Yim, Hwang, Yun, Han, and Park]{kim2021donut}
Geewook Kim, Teakgyu Hong, Moonbin Yim, Jinyoung Park, Jinyeong Yim, Wonseok Hwang, Sangdoo Yun, Dongyoon Han, and Seunghyun Park.
\newblock Donut: Document understanding transformer without ocr.
\newblock \emph{arXiv preprint arXiv:2111.15664}, 7\penalty0 (15):\penalty0 2, 2021.

\bibitem[Lauren{\c{c}}on et~al.(2024)Lauren{\c{c}}on, Tronchon, Cord, and Sanh]{laurenccon2024matters}
Hugo Lauren{\c{c}}on, L{\'e}o Tronchon, Matthieu Cord, and Victor Sanh.
\newblock What matters when building vision-language models?
\newblock \emph{arXiv preprint arXiv:2405.02246}, 2024.

\bibitem[Li et~al.(2024{\natexlab{a}})Li, Zhang, Guo, Zhang, Li, Zhang, Zhang, Li, Liu, and Li]{li2024llava}
Bo Li, Yuanhan Zhang, Dong Guo, Renrui Zhang, Feng Li, Hao Zhang, Kaichen Zhang, Yanwei Li, Ziwei Liu, and Chunyuan Li.
\newblock Llava-onevision: Easy visual task transfer.
\newblock \emph{arXiv preprint arXiv:2408.03326}, 2024{\natexlab{a}}.

\bibitem[Li et~al.(2023)Li, Li, Savarese, and Hoi]{li2023blip}
Junnan Li, Dongxu Li, Silvio Savarese, and Steven Hoi.
\newblock Blip-2: Bootstrapping language-image pre-training with frozen image encoders and large language models.
\newblock In \emph{International conference on machine learning}, pages 19730--19742. PMLR, 2023.

\bibitem[Li et~al.(2024{\natexlab{b}})Li, Yuan, Liu, Tang, Wang, Zhu, and Zhang]{li2024tokenpacker}
Wentong Li, Yuqian Yuan, Jian Liu, Dongqi Tang, Song Wang, Jianke Zhu, and Lei Zhang.
\newblock Tokenpacker: Efficient visual projector for multimodal llm.
\newblock \emph{arXiv preprint arXiv:2407.02392}, 2024{\natexlab{b}}.

\bibitem[Litman et~al.(2020)Litman, Anschel, Tsiper, Litman, Mazor, and Manmatha]{litman2020scatter}
Ron Litman, Oron Anschel, Shahar Tsiper, Roee Litman, Shai Mazor, and R Manmatha.
\newblock Scatter: selective context attentional scene text recognizer.
\newblock In \emph{proceedings of the IEEE/CVF conference on computer vision and pattern recognition}, pages 11962--11972, 2020.

\bibitem[Liu et~al.(2024{\natexlab{a}})Liu, Li, Li, Li, Zhang, Shen, and Lee]{liu2024llavanext}
Haotian Liu, Chunyuan Li, Yuheng Li, Bo Li, Yuanhan Zhang, Sheng Shen, and Yong~Jae Lee.
\newblock Llava-next: Improved reasoning, ocr, and world knowledge, 2024{\natexlab{a}}.

\bibitem[Liu et~al.(2024{\natexlab{b}})Liu, Li, Wu, and Lee]{liu2024visual}
Haotian Liu, Chunyuan Li, Qingyang Wu, and Yong~Jae Lee.
\newblock Visual instruction tuning.
\newblock \emph{Advances in neural information processing systems}, 36, 2024{\natexlab{b}}.

\bibitem[Liu et~al.(2024{\natexlab{c}})Liu, Li, Huang, Yang, Yu, Li, Yin, lin Liu, Jin, and Bai]{liu2024ocrbenchhiddenmysteryocr}
Yuliang Liu, Zhang Li, Mingxin Huang, Biao Yang, Wenwen Yu, Chunyuan Li, Xucheng Yin, Cheng lin Liu, Lianwen Jin, and Xiang Bai.
\newblock Ocrbench: On the hidden mystery of ocr in large multimodal models, 2024{\natexlab{c}}.

\bibitem[Masry et~al.(2022)Masry, Long, Tan, Joty, and Hoque]{masry2022chartqa}
Ahmed Masry, Do~Xuan Long, Jia~Qing Tan, Shafiq Joty, and Enamul Hoque.
\newblock Chartqa: A benchmark for question answering about charts with visual and logical reasoning.
\newblock \emph{arXiv preprint arXiv:2203.10244}, 2022.

\bibitem[Mathew et~al.(2021)Mathew, Karatzas, and Jawahar]{mathew2021docvqa}
Minesh Mathew, Dimosthenis Karatzas, and CV Jawahar.
\newblock Docvqa: A dataset for vqa on document images.
\newblock In \emph{Proceedings of the IEEE/CVF winter conference on applications of computer vision}, pages 2200--2209, 2021.

\bibitem[Mathew et~al.(2022)Mathew, Bagal, Tito, Karatzas, Valveny, and Jawahar]{mathew2022infographicvqa}
Minesh Mathew, Viraj Bagal, Rub{\`e}n Tito, Dimosthenis Karatzas, Ernest Valveny, and CV Jawahar.
\newblock Infographicvqa.
\newblock In \emph{Proceedings of the IEEE/CVF Winter Conference on Applications of Computer Vision}, pages 1697--1706, 2022.

\bibitem[Mishra et~al.(2019)Mishra, Shekhar, Singh, and Chakraborty]{mishra2019ocr}
Anand Mishra, Shashank Shekhar, Ajeet~Kumar Singh, and Anirban Chakraborty.
\newblock Ocr-vqa: Visual question answering by reading text in images.
\newblock In \emph{2019 international conference on document analysis and recognition (ICDAR)}, pages 947--952. IEEE, 2019.

\bibitem[Nuriel et~al.(2022)Nuriel, Fogel, and Litman]{nuriel2022textadain}
Oren Nuriel, Sharon Fogel, and Ron Litman.
\newblock Textadain: Paying attention to shortcut learning in text recognizers.
\newblock In \emph{European Conference on Computer Vision}, pages 427--445. Springer, 2022.

\bibitem[Peng et~al.(2023)Peng, Wang, Dong, Hao, Huang, Ma, and Wei]{peng2023kosmos}
Zhiliang Peng, Wenhui Wang, Li Dong, Yaru Hao, Shaohan Huang, Shuming Ma, and Furu Wei.
\newblock Kosmos-2: Grounding multimodal large language models to the world.
\newblock \emph{arXiv preprint arXiv:2306.14824}, 2023.

\bibitem[Raffel et~al.(2020)Raffel, Shazeer, Roberts, Lee, Narang, Matena, Zhou, Li, and Liu]{raffel2020exploring}
Colin Raffel, Noam Shazeer, Adam Roberts, Katherine Lee, Sharan Narang, Michael Matena, Yanqi Zhou, Wei Li, and Peter~J Liu.
\newblock Exploring the limits of transfer learning with a unified text-to-text transformer.
\newblock \emph{Journal of machine learning research}, 21\penalty0 (140):\penalty0 1--67, 2020.

\bibitem[Ronen et~al.(2022)Ronen, Tsiper, Anschel, Lavi, Markovitz, and Manmatha]{ronen2022glass}
Roi Ronen, Shahar Tsiper, Oron Anschel, Inbal Lavi, Amir Markovitz, and R Manmatha.
\newblock Glass: Global to local attention for scene-text spotting.
\newblock \emph{arXiv preprint arXiv:2208.03364}, 2022.

\bibitem[Sidorov et~al.(2020)Sidorov, Hu, Rohrbach, and Singh]{sidorov2020textcaps}
Oleksii Sidorov, Ronghang Hu, Marcus Rohrbach, and Amanpreet Singh.
\newblock Textcaps: a dataset for image captioning with reading comprehension.
\newblock In \emph{Computer Vision--ECCV 2020: 16th European Conference, Glasgow, UK, August 23--28, 2020, Proceedings, Part II 16}, pages 742--758. Springer, 2020.

\bibitem[Singh et~al.(2019)Singh, Natarajan, Shah, Jiang, Chen, Batra, Parikh, and Rohrbach]{singh2019towards}
Amanpreet Singh, Vivek Natarajan, Meet Shah, Yu Jiang, Xinlei Chen, Dhruv Batra, Devi Parikh, and Marcus Rohrbach.
\newblock Towards vqa models that can read.
\newblock In \emph{Proceedings of the IEEE/CVF conference on computer vision and pattern recognition}, pages 8317--8326, 2019.

\bibitem[Tanaka et~al.(2024)Tanaka, Iki, Nishida, Saito, and Suzuki]{tanaka2024instructdoc}
Ryota Tanaka, Taichi Iki, Kyosuke Nishida, Kuniko Saito, and Jun Suzuki.
\newblock Instructdoc: A dataset for zero-shot generalization of visual document understanding with instructions.
\newblock In \emph{Proceedings of the AAAI Conference on Artificial Intelligence}, pages 19071--19079, 2024.

\bibitem[Tito et~al.(2023)Tito, Karatzas, and Valveny]{tito2023hierarchical}
Rub{\`e}n Tito, Dimosthenis Karatzas, and Ernest Valveny.
\newblock Hierarchical multimodal transformers for multipage docvqa.
\newblock \emph{Pattern Recognition}, 144:\penalty0 109834, 2023.

\bibitem[Van~Landeghem et~al.(2023)Van~Landeghem, Tito, Borchmann, Pietruszka, Joziak, Powalski, Jurkiewicz, Coustaty, Anckaert, Valveny, et~al.]{van2023document}
Jordy Van~Landeghem, Rub{\`e}n Tito, {\L}ukasz Borchmann, Micha{\l} Pietruszka, Pawel Joziak, Rafal Powalski, Dawid Jurkiewicz, Micka{\"e}l Coustaty, Bertrand Anckaert, Ernest Valveny, et~al.
\newblock Document understanding dataset and evaluation (dude).
\newblock In \emph{Proceedings of the IEEE/CVF International Conference on Computer Vision}, pages 19528--19540, 2023.

\bibitem[Wang et~al.(2023{\natexlab{a}})Wang, Raman, Sibue, Ma, Babkin, Kaur, Pei, Nourbakhsh, and Liu]{wang2023docllm}
Dongsheng Wang, Natraj Raman, Mathieu Sibue, Zhiqiang Ma, Petr Babkin, Simerjot Kaur, Yulong Pei, Armineh Nourbakhsh, and Xiaomo Liu.
\newblock Docllm: A layout-aware generative language model for multimodal document understanding.
\newblock \emph{arXiv preprint arXiv:2401.00908}, 2023{\natexlab{a}}.

\bibitem[Wang et~al.(2024)Wang, Bai, Tan, Wang, Fan, Bai, Chen, Liu, Wang, Ge, et~al.]{wang2024qwen2}
Peng Wang, Shuai Bai, Sinan Tan, Shijie Wang, Zhihao Fan, Jinze Bai, Keqin Chen, Xuejing Liu, Jialin Wang, Wenbin Ge, et~al.
\newblock Qwen2-vl: Enhancing vision-language model's perception of the world at any resolution.
\newblock \emph{arXiv preprint arXiv:2409.12191}, 2024.

\bibitem[Wang et~al.(2023{\natexlab{b}})Wang, Li, Ou, and Zhang]{wang2023layout}
Wenjin Wang, Yunhao Li, Yixin Ou, and Yin Zhang.
\newblock Layout and task aware instruction prompt for zero-shot document image question answering.
\newblock \emph{arXiv preprint arXiv:2306.00526}, 2023{\natexlab{b}}.

\bibitem[Xu et~al.(2020)Xu, Li, Cui, Huang, Wei, and Zhou]{xu2020layoutlm}
Yiheng Xu, Minghao Li, Lei Cui, Shaohan Huang, Furu Wei, and Ming Zhou.
\newblock Layoutlm: Pre-training of text and layout for document image understanding.
\newblock In \emph{Proceedings of the 26th ACM SIGKDD international conference on knowledge discovery \& data mining}, pages 1192--1200, 2020.

\bibitem[Ye et~al.(2023{\natexlab{a}})Ye, Hu, Xu, Ye, Yan, Xu, Li, Tian, Qian, Zhang, et~al.]{ye2023ureader}
Jiabo Ye, Anwen Hu, Haiyang Xu, Qinghao Ye, Ming Yan, Guohai Xu, Chenliang Li, Junfeng Tian, Qi Qian, Ji Zhang, et~al.
\newblock Ureader: Universal ocr-free visually-situated language understanding with multimodal large language model.
\newblock \emph{arXiv preprint arXiv:2310.05126}, 2023{\natexlab{a}}.

\bibitem[Ye et~al.(2023{\natexlab{b}})Ye, Zhang, Zhao, Liu, Du, and Tao]{ye2022dptext}
Maoyuan Ye, Jing Zhang, Shanshan Zhao, Juhua Liu, Bo Du, and Dacheng Tao.
\newblock Dptext-detr: Towards better scene text detection with dynamic points in transformer.
\newblock In \emph{Proceedings of the AAAI Conference on Artificial Intelligence}, pages 3241--3249, 2023{\natexlab{b}}.

\bibitem[Ye et~al.(2023{\natexlab{c}})Ye, Zhang, Zhao, Liu, Liu, Du, and Tao]{ye2023deepsolo}
Maoyuan Ye, Jing Zhang, Shanshan Zhao, Juhua Liu, Tongliang Liu, Bo Du, and Dacheng Tao.
\newblock Deepsolo: Let transformer decoder with explicit points solo for text spotting.
\newblock In \emph{Proceedings of the IEEE/CVF Conference on Computer Vision and Pattern Recognition}, pages 19348--19357, 2023{\natexlab{c}}.

\bibitem[Ye et~al.(2023{\natexlab{d}})Ye, Xu, Xu, Ye, Yan, Zhou, Wang, Hu, Shi, Shi, et~al.]{ye2023mplug}
Qinghao Ye, Haiyang Xu, Guohai Xu, Jiabo Ye, Ming Yan, Yiyang Zhou, Junyang Wang, Anwen Hu, Pengcheng Shi, Yaya Shi, et~al.
\newblock mplug-owl: Modularization empowers large language models with multimodality.
\newblock \emph{arXiv preprint arXiv:2304.14178}, 2023{\natexlab{d}}.

\bibitem[Zhang et~al.(2024)Zhang, Dong, Zang, Cao, Qian, Chen, Guo, Duan, Wang, Ouyang, et~al.]{zhang2024internlm}
Pan Zhang, Xiaoyi Dong, Yuhang Zang, Yuhang Cao, Rui Qian, Lin Chen, Qipeng Guo, Haodong Duan, Bin Wang, Linke Ouyang, et~al.
\newblock Internlm-xcomposer-2.5: A versatile large vision language model supporting long-contextual input and output.
\newblock \emph{arXiv preprint arXiv:2407.03320}, 2024.

\bibitem[Zhu et~al.(2023)Zhu, Chen, Shen, Li, and Elhoseiny]{zhu2023minigpt}
Deyao Zhu, Jun Chen, Xiaoqian Shen, Xiang Li, and Mohamed Elhoseiny.
\newblock Minigpt-4: Enhancing vision-language understanding with advanced large language models.
\newblock \emph{arXiv preprint arXiv:2304.10592}, 2023.

\bibitem[Zhu et~al.(2022)Zhu, Lei, Feng, Wang, Zhang, and Chua]{zhu2022towards}
Fengbin Zhu, Wenqiang Lei, Fuli Feng, Chao Wang, Haozhou Zhang, and Tat-Seng Chua.
\newblock Towards complex document understanding by discrete reasoning.
\newblock In \emph{Proceedings of the 30th ACM International Conference on Multimedia}, pages 4857--4866, 2022.

\end{thebibliography}
}

\clearpage

\appendix
\maketitlesupplementary

\section{Additional Implementation Details} \label{sup sec:Implementation details}

DocVLM efficiently leverages OCR data to enhance VLMs' reading capabilities. 
We extract text and layout information from document images using an OCR system, which is then processed by an OCR encoder as discussed in Sec. \refwithdefault{sec: method}{3}.
Specifically, we utilize the encoder component of DocFormerV2~\cite{appalaraju2024docformerv2}, omitting the visual branch of this encoder, as detailed in the main paper. The encoder is initialized with pretrained weights from DocFormerV2, which was pretrained on the Industry Document Library (IDL) dataset~\cite{biten2022ocr}. For details on this pretraining process, refer to \cite{appalaraju2024docformerv2}.

\subsection{Optimization and Hyperparameters Details}

As discussed in Sec.~\refwithdefault{sec:training}{3.2}, our training process comprises two stages: 1) OCR-LLM alignment and 2) Vision alignment. For both stages, we utilize AdamW optimization algorithm with a cosine learning scheduling and $1000$ warmup steps.
For the OCR-LLM alignment stage with our learned queries component, we trained for $140$K steps. We used learning rates of $10^{-4}$ for the projection layer and query tokens, and $5\cdot10^{-5}$ for the OCR encoder. To preserve the pretrained weights of the OCR encoder while optimizing the randomly initialized components, we initially froze the encoder for the first $10$K steps. In experiments without our learned queries component (\textit{i.e.}, without OCR compression), we adjusted the process, training the OCR encoder and projection layer for $100$K steps using the same learning rates.
In the Vision alignment stage, we trained all components for an additional $100$K steps with a learning rate of $5\cdot10^{-6}$. Unlike the previous stage, this phase included visual features as input to the LLM, allowing the model to align the OCR modality with the visual one.

\section{Datasets}
\subsection{Training Datasets}

\cref{tab:training_datasets} details all the datasets used to fine-tune DocVLM. For the OCR-LLM alignment stage, our dataset selection focuses on text-related tasks, including approximately $990$K queries, including document VQA datasets (DocVQA~\cite{mathew2021docvqa}, InfoVQA~\cite{mathew2022infographicvqa}, ChartQA~\cite{masry2022chartqa}, TAT-DQA~\cite{zhu2022towards}), scene text VQA datasets (TextVQA~\cite{singh2019towards}, ST-VQA~\cite{biten2019scene}, OCR-VQA~\cite{mishra2019ocr}), and a captioning dataset (TextCaps~\cite{sidorov2020textcaps}). The vision alignment stage incorporates additional visual-centric datasets: COCO Caption~\cite{chen2015microsoft} and VQA-V2~\cite{balanced_vqa_v2}, bringing the total training set to approximately $2$M queries.

\begin{table}[t]
\centering
\resizebox{\linewidth}{!}{
\begin{tabular}{lllcc}
\toprule
\textbf{Task} & \textbf{Dataset} & \textbf{Subsplit} & \textbf{Visual Only}  & \textbf{\# Queries} \\
\midrule
\multirow{4}{*}{Document VQA} & DocVQA~\cite{mathew2021docvqa}  & train    & $\times$ & 39463 \\
                             & InfoVQA~\cite{mathew2022infographicvqa}  & train   & $\times$ & 46883\\
                             & ChartQA ~\cite{masry2022chartqa}    & train (H) & $\times$ & 7398\\
                             & TAT-DQA~\cite{zhu2022towards}    & train & $\times$ & 13246\\
\midrule
\multirow{3}{*}{Scene Text VQA}  & TextVQA~\cite{singh2019towards}  & train   & $\times$ & 34602 \\
                             & ST-VQA~\cite{biten2019scene}   & train & $\times$ & 26308\\
                             & OCR-VQA~\cite{mishra2019ocr}   & train  & $\times$ & 800000 \\
\midrule
\multirow{2}{*}{Captioning}  & TextCaps~\cite{sidorov2020textcaps}   & train & $\times$ & 21953\\
                            & COCO Caption~\cite{chen2015microsoft}  & train & $\checkmark$ & 566747\\
\midrule
General VQA & VQA-V2~\cite{balanced_vqa_v2}  & train  & $\checkmark$ & 443757\\
\midrule
\textbf{Total Examples} & & &  & \textbf{2000357} \\
\bottomrule
\end{tabular}}
\caption{\textbf{Training Datasets for DocVLM Fine-tuning.} Datasets used for fine-tuning DocVLM, categorized by task type. The 'Visual Only' column indicates datasets that are not text-centric. The total number of queries across all datasets is shown at the bottom.} \label{tab:training_datasets}
\hspace{0.5em}
\end{table}

\subsection{Evaluation Datasets}

\cref{tab:eval_datasets} details all the datasets used to evaluate DocVLM's performance across a diverse range of document understanding tasks, including document VQA, scene text VQA, captioning, and multipage document understanding. While our training focused on single-page documents, we extended our evaluation to include multipage datasets: MP-DocVQA~\cite{tito2023hierarchical} and DUDE~\cite{van2023document}. It is important to note that although both multipage datasets were not included in our training set, we only consider DUDE as a true zero-shot evaluation, as MP-DocVQA is an extension of DocVQA, which was included in our training data.

\begin{table}[t]
\centering
\resizebox{\linewidth}{!}{
\begin{tabular}{llllcc}
\toprule
\textbf{Task} & \textbf{Dataset} & \textbf{Subsplit} & \textbf{Metric} & \textbf{Zero-Shot}  &\textbf{\# Queries} \\
\midrule
\multirow{2}{*}{Document VQA} & DocVQA~\cite{mathew2021docvqa}     & Test & ANLS & $\times$  & 5188\\
                             & InfoVQA~\cite{mathew2022infographicvqa}    & Test & ANLS & $\times$ & 6573\\
\midrule
\multirow{2}{*}{Scene Text VQA}  & TextVQA~\cite{singh2019towards}    & Val & VQAScore & $\times$  & 5000\\
                             & ST-VQA~\cite{biten2019scene}    & Test & ANLS & $\times$ & 4163 \\
\midrule
\multirow{1}{*}{Captioning}  & TextCaps~\cite{sidorov2020textcaps}   & Val & CIDEr & $\times$  & 3166\\
\midrule
\multirow{2}{*}{Multipage VQA}  & MP-DocVQA~\cite{tito2023hierarchical}   & Test & ANLS & $\times$  & 5019\\
                            & DUDE~\cite{van2023document}   & Test & ANLS & $\checkmark$ & 11402\\
\midrule
\textbf{Total Examples} & &  & & & \textbf{40511} \\
\bottomrule
\end{tabular}}
\caption{\textbf{Evaluation Datasets for DocVLM.} Datasets used for evaluating DocVLM, categorized by task type. The table includes the dataset split used, evaluation metric, zero-shot status, and number of queries for each dataset.} \label{tab:eval_datasets}
\end{table}

\newpage

\section{Additional Results} \label{sup sec:Additional Experiments}

\subsection{Qualitative Results}

\begin{figure*}[h]
    \raggedright
    
    \begin{subfigure}[b]{0.45\textwidth}
        \centering
        \includegraphics[height=5.6cm]{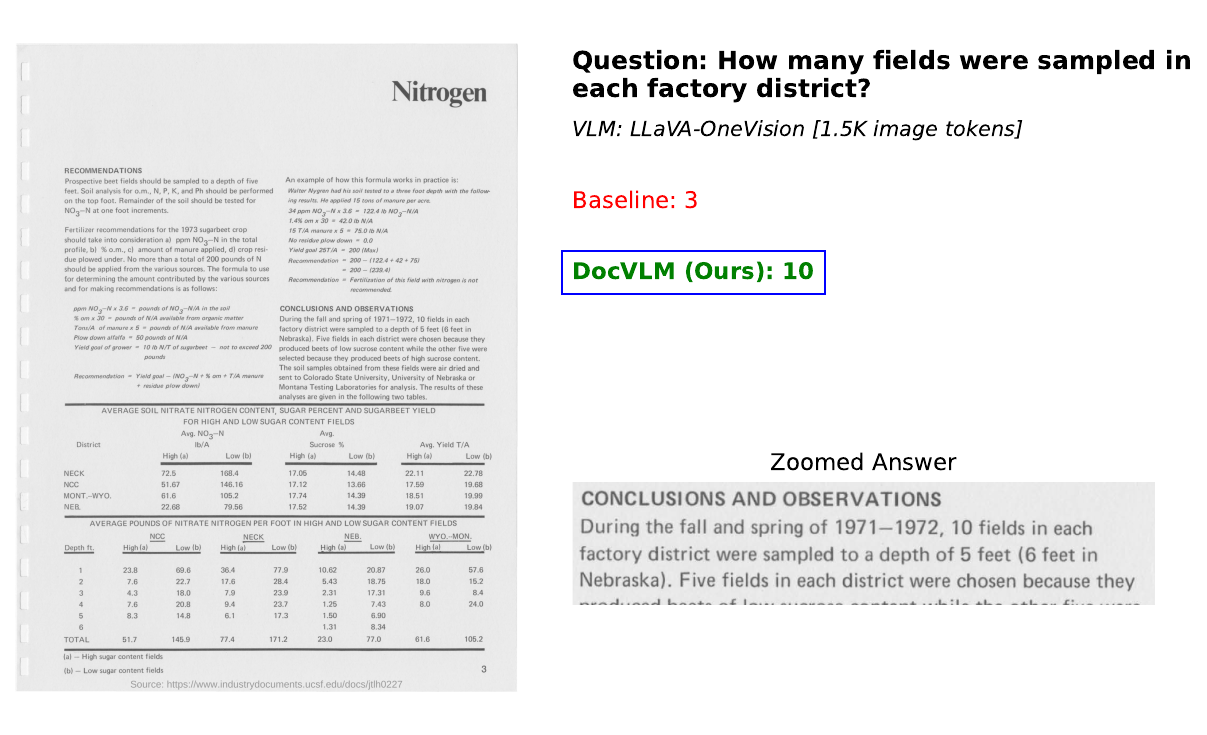}
    \end{subfigure}
    \hfill
    \begin{subfigure}[b]{0.45\textwidth}
        \centering
        \includegraphics[height=5.6cm, trim=1.5cm 0 0 0, clip]{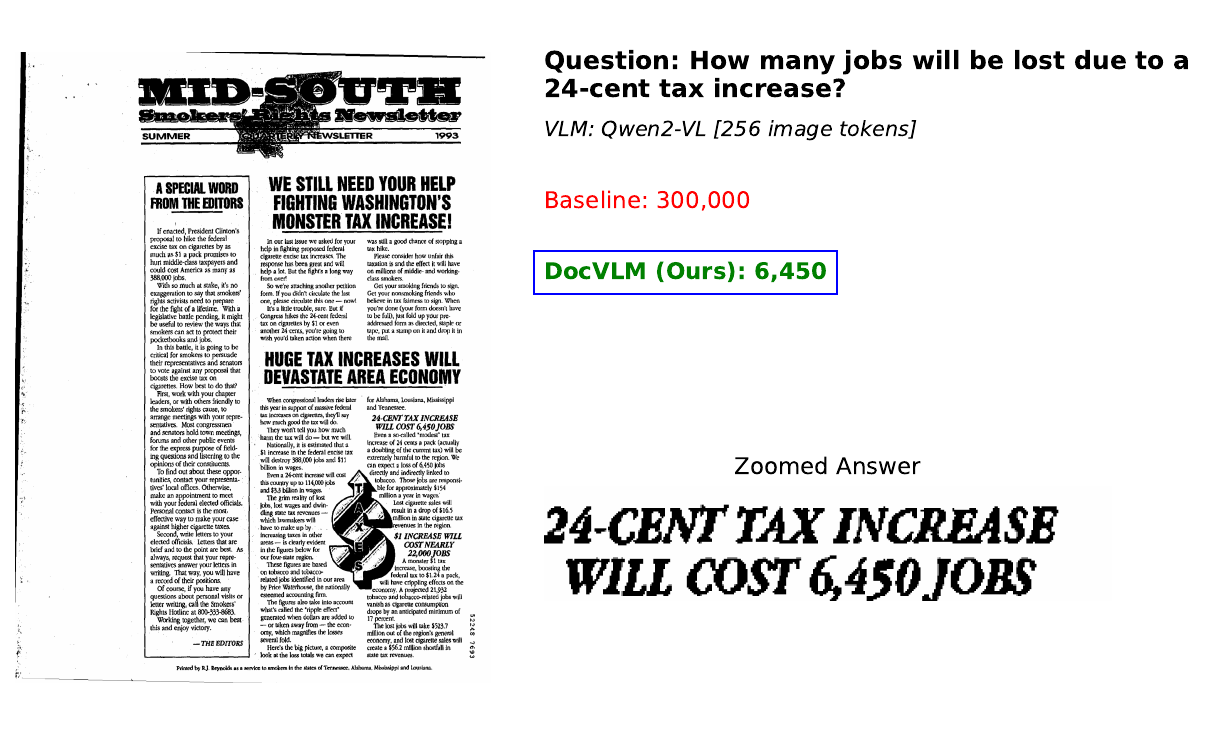}
    \end{subfigure}
    \vfill    
    \begin{subfigure}[b]{0.45\textwidth}
        \centering
        \includegraphics[height=5.6cm]{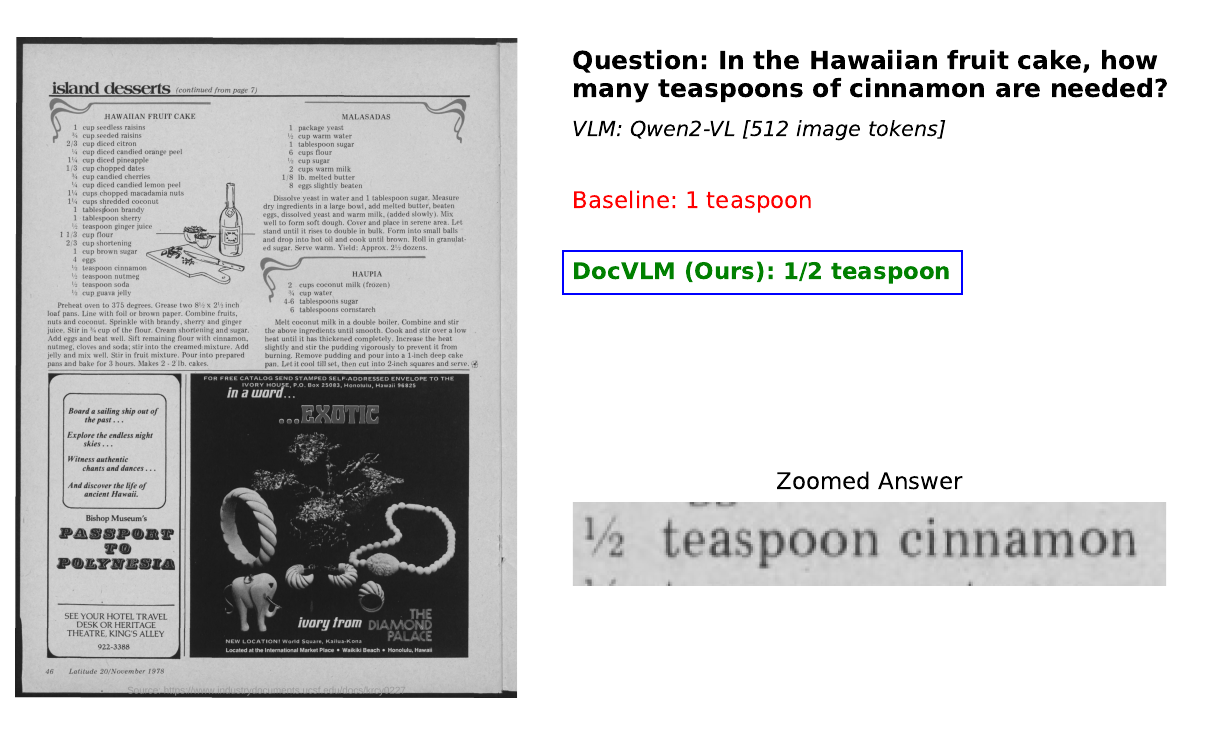}
    \end{subfigure}
    \hfill
    \begin{subfigure}[b]{0.45\textwidth}
        \centering
        \includegraphics[height=5.6cm]{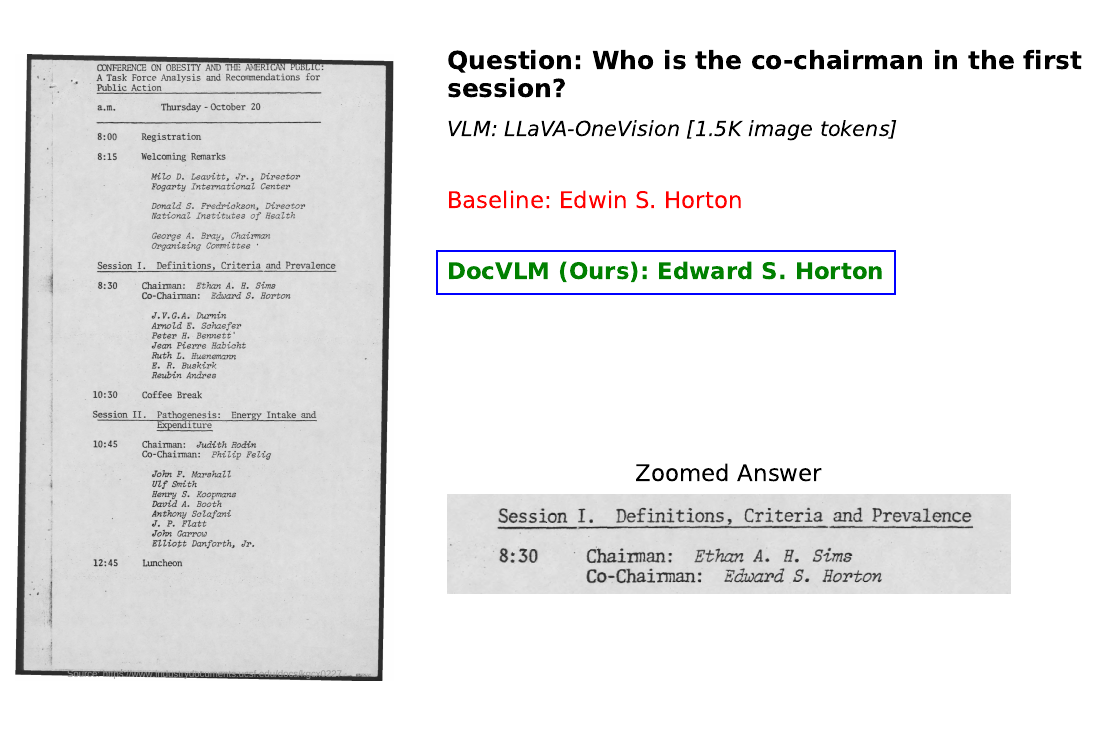}
    \end{subfigure}
    \vfill
    \begin{subfigure}[b]{0.45\textwidth}
        \centering
        \includegraphics[height=5.6cm]{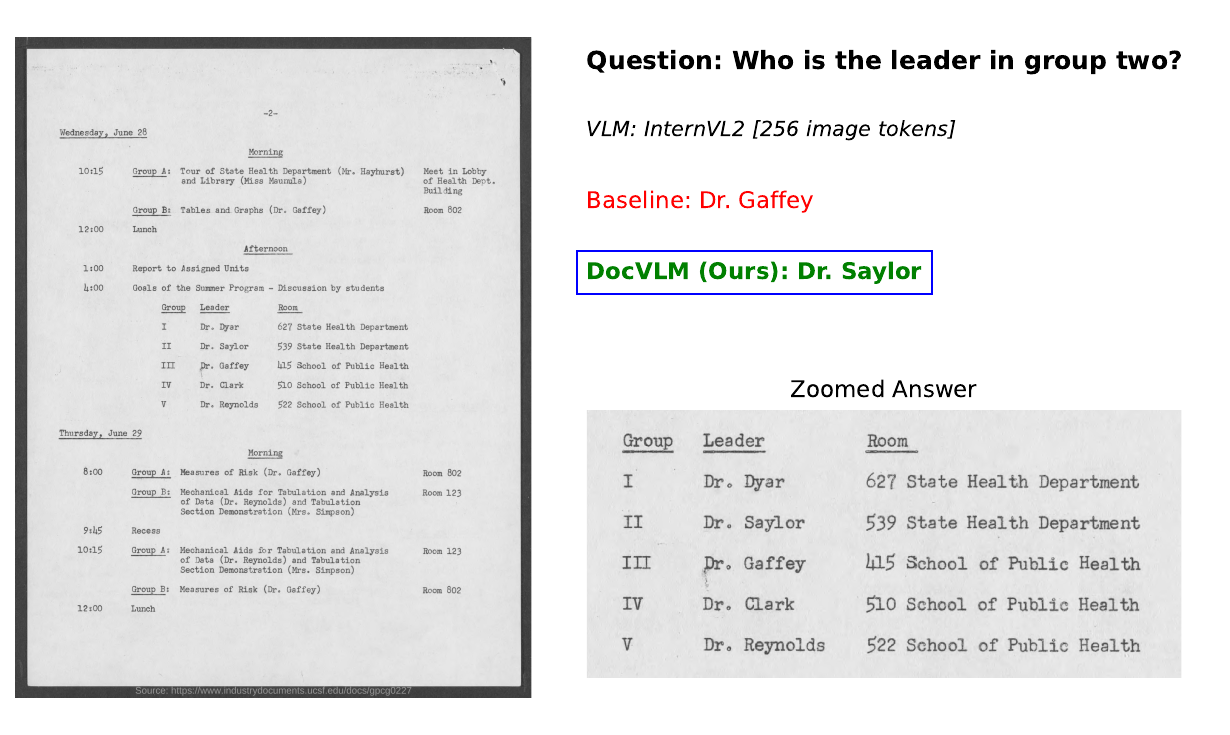}
    \end{subfigure}
    \hfill
    \begin{subfigure}[b]{0.45\textwidth}
        \centering
        \includegraphics[height=5.6cm, trim=1.5cm 0 0 0, clip]{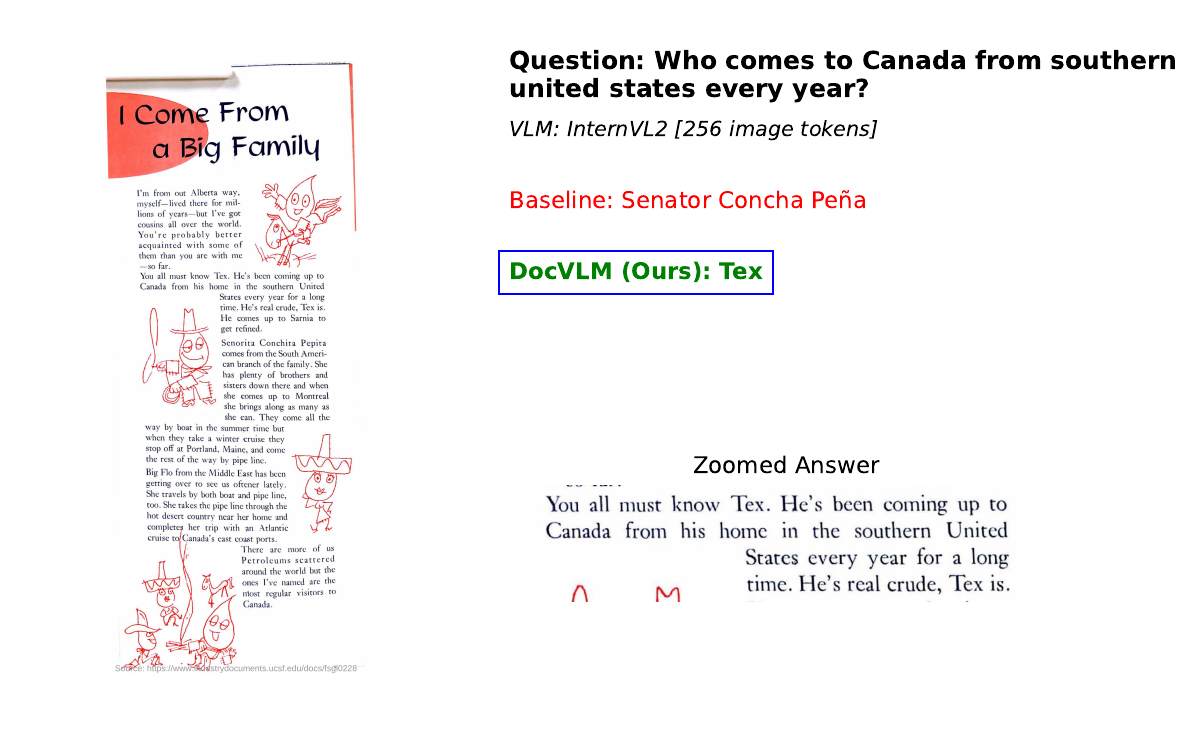}
    \end{subfigure}
    \caption{\textbf{Qualitative Results on Text-Heavy Documents.} Representative examples of DocVLM's performance on text-dense documents compared to baseline models (LLaVA-OneVision, InternVL2, and Qwen2VL). Each example shows an image-instruction pair with baseline and DocVLM predictions, demonstrating DocVLM's enhanced reading comprehension using only 64 OCR compressed tokens.}
\label{fig:sup_qualitative_1}
\end{figure*}

\begin{figure*}[h]
    \raggedright

    \begin{subfigure}[b]{0.6\textwidth}
        \centering
        \includegraphics[height=5.3cm]{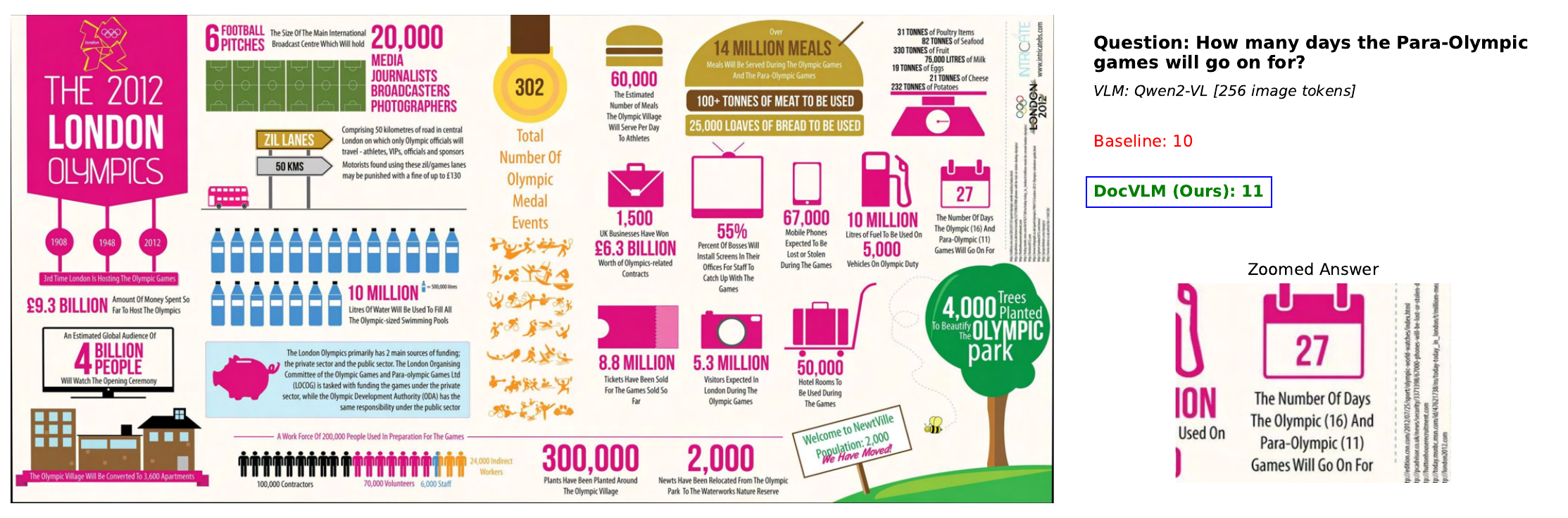}
    \end{subfigure}
    \vfill
    \begin{subfigure}[b]{0.6\textwidth}
        \centering
        \includegraphics[height=5.3cm]{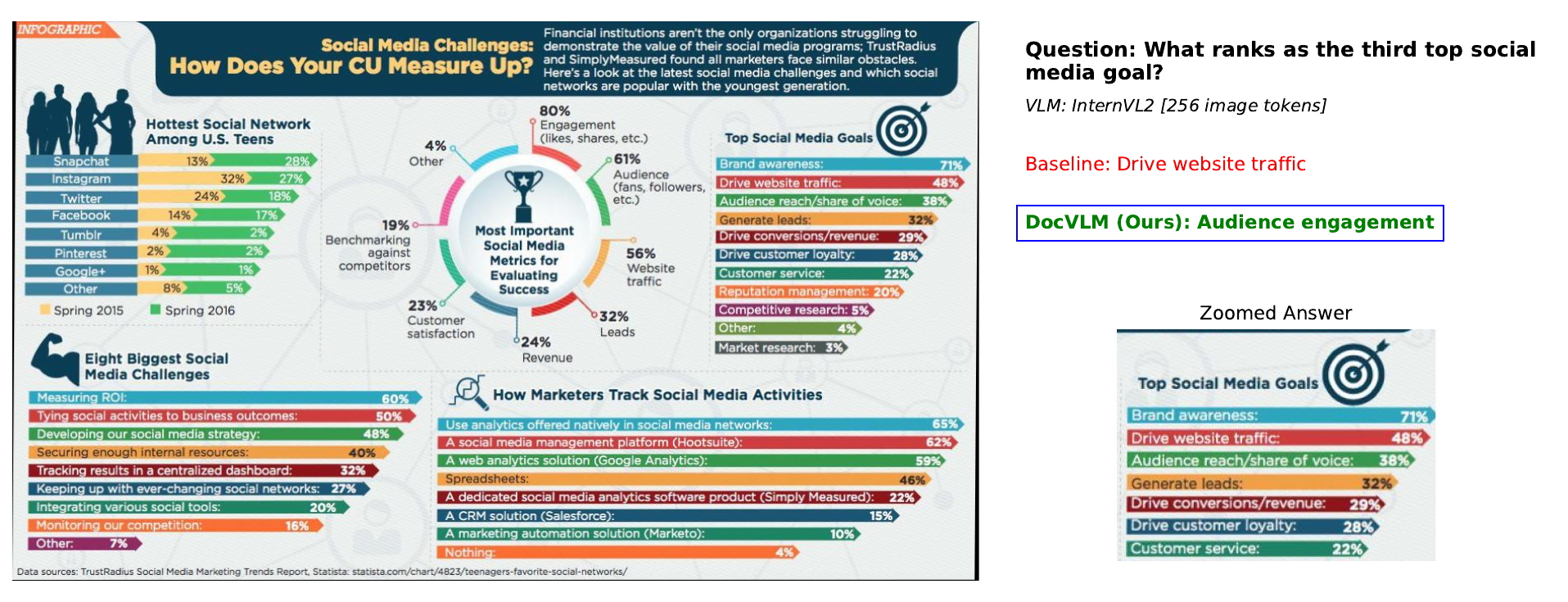}
    \end{subfigure}
    \vfill
    \begin{subfigure}[b]{0.45\textwidth}
        \centering
        \includegraphics[height=5.3cm]{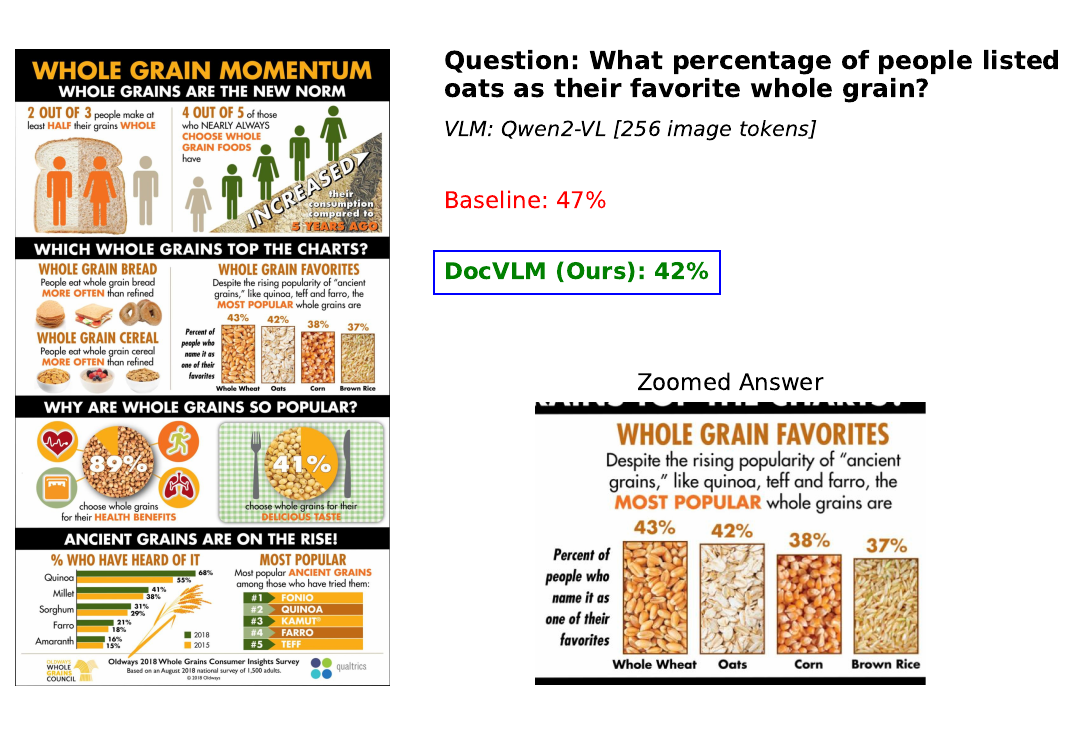}
    \end{subfigure}
    \hspace{1.4em}
    \begin{subfigure}[b]{0.45\textwidth}
        \centering
        \includegraphics[height=5.3cm]{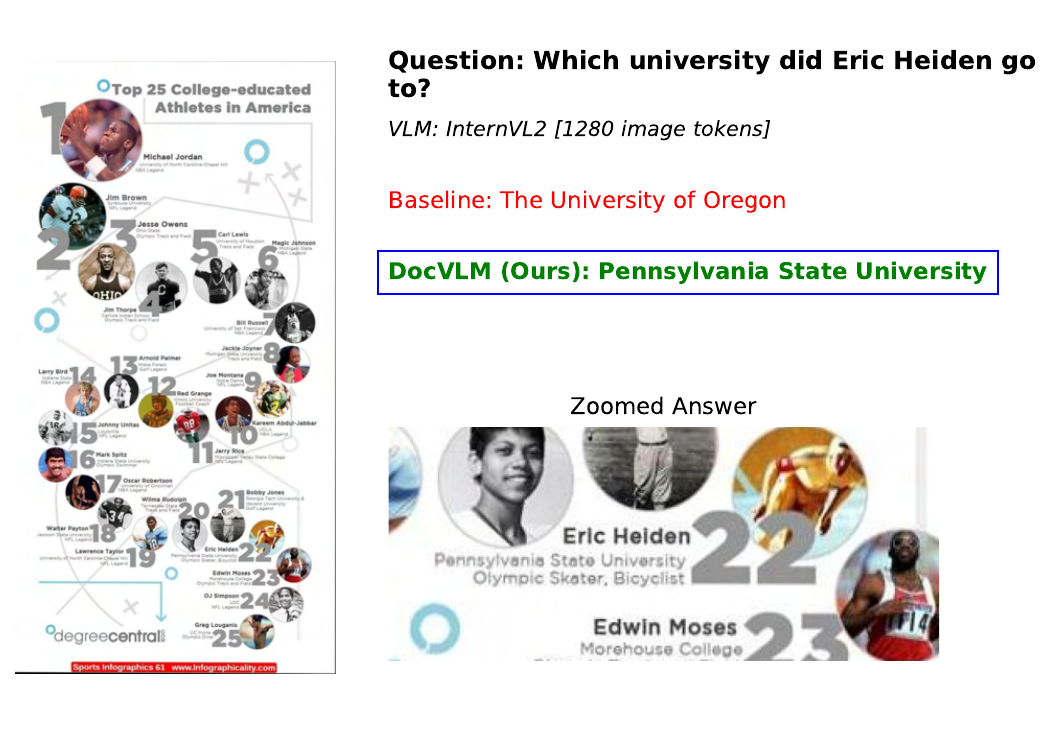}
    \end{subfigure}
    \vfill
    \begin{subfigure}[b]{0.6\textwidth}
        \centering
        \includegraphics[height=5.3cm]{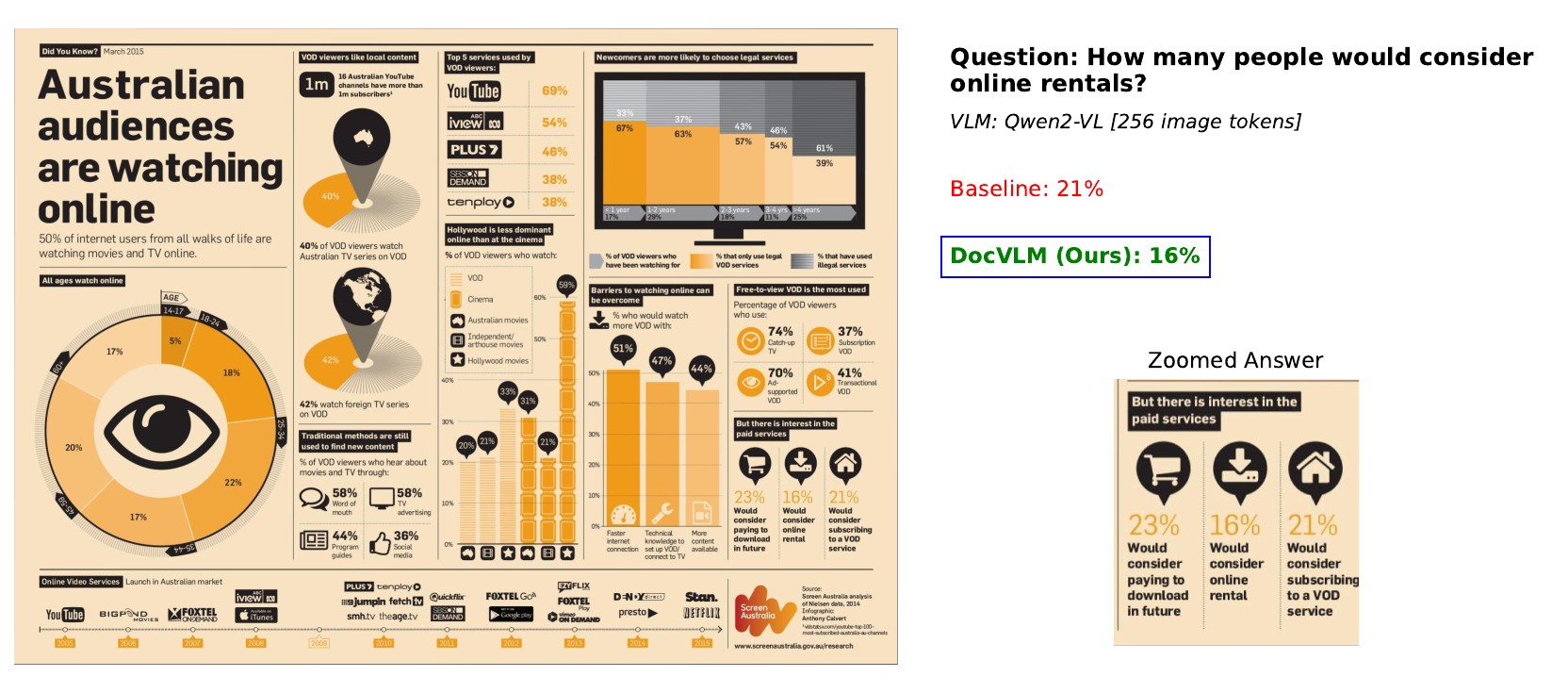}
    \end{subfigure}
    \caption{\textbf{Qualitative Results on Infographics.}  Representative examples of  DocVLM's performance on infographic-style documents compared to baselines under various visual token constraints, demonstrating improved handling of complex layouts and visual information.}
    \label{fig:sup_qualitative_2}
\end{figure*}

Figures \ref{fig:sup_qualitative_1} and \ref{fig:sup_qualitative_2} showcase DocVLM's enhanced document understanding capabilities through representative examples. Figure \ref{fig:sup_qualitative_1} focuses on document images from the DocVQA~\cite{mathew2021docvqa} test set, while Figure \ref{fig:sup_qualitative_2} presents infographic images from the InfoVQA test set~\cite{mathew2022infographicvqa}. We present results for LLaVA-OneVision with a $1.5$K visual token limitation, InternVL2 with $256$ and $1280$ visual token limitations, and Qwen2VL with $256$ and $512$ visual token limitations. 
As can be seen, the baselines' errors occur in scenarios that demand superior reading comprehension capabilities.
Notably, by only utilizing $64$ OCR compressed tokens, DocVLM effectively corrects errors and provides the correct responses.
This improvement is consistent across different VLM architectures and visual token limitations, highlighting the efficiency and versatility of our approach.

\section{Studying The Visual Features Effect}

In this section, we explore how visual features contribute to DocVLM's performance by first evaluating DocVLM \emph{without visual input} and then assessing the impact of adding visual features.

\paragraph{DocVLM's OCR Encodings Without Visual Input.} We evaluate DocVLM based on Qwen2VL after the OCR-LLM Alignment stage, using only OCR encodings as input to the LLM, without visual tokens. This approach allows us to assess how well the encodings capture OCR data and their sufficiency for document question answering tasks. Our architecture consists of inputting DocVLM's encodings or compressed encodings to the Qwen2 LLM along with the query prompt. \cref{tab: DocVLM_encoding_in_llm} presents our results on DocVQA~\cite{mathew2021docvqa} and InfoVQA~\cite{mathew2022infographicvqa} test sets compared to baselines that also rely solely on OCR information~\cite{wang2023layout, xu2020layoutlm, wang2023docllm}. We can see that DocVLM's OCR encodings effectively capture OCR information, yielding the best results in the comparison. Remarkably, using only 64 learned queries (compressed encodings) achieves competitive performance, significantly surpassing the OCR words baseline, despite being much shorter (64 compared to ~1K tokens).

\begin{table}[ht]
\centering
\resizebox{\linewidth}{!}{
\begin{tabular}{llcc}
\toprule
\textbf{Method} & \textbf{LLM OCR Input} & {DocVQA} & InfoVQA \\
\midrule
Alpaca & Latin Prompt  & 42.0 & -- \\
ChatGPT-3.5 & Latin Prompt  & 82.6 & 49.0 \\
LayoutLM$_{\text{LARGE}}$ & OCR Encodings &  72.6 & 27.2 \\
DocLLM & OCR Encodings  & 69.5 & -- \\
\hdashline
Qwen2 & OCR Words & 76.4 & 44.5 \\
\rowcolor{cyan!10} \textbf{$\text{DocVLM}_{\text{Qwen2}}$} & OCR Encodings   & \textbf{89.2} & \textbf{62.9} \\
\rowcolor{cyan!10} \textbf{$\text{DocVLM}_{\text{Qwen2}}$} & 64 Compressed Encodings   & \underline{85.5} & \underline{56.8} \\
\bottomrule
\end{tabular}}
\caption{\textbf{Effectiveness in LLMs (no visual input).} Comparison of DocVLM's full and compressed OCR encodings as \emph{sole} input to Qwen2 LLM against OCR-only baselines, showing DocVLM's OCR encodings effectiveness even without visual features.} \label{tab: DocVLM_encoding_in_llm}
\vspace{-0.1cm}
\end{table}


\paragraph{Contribution of Visual Features.} In \cref{tab: visual features effect}, we compare the results from the previous text-only evaluation to those obtained when adding $256$ visual tokens to the input of the same model checkpoint. The results demonstrate that incorporating visual information improves performance across both datasets, with a particularly notable enhancement when using compressed OCR encodings. This comparison highlights the complementary nature of textual and visual information in DocVLM's architecture.
\begin{table}[ht]
\centering
\resizebox{\linewidth}{!}{
\begin{tabular}{c c c c c} 
\toprule 
\multirow{2}{*}{\textbf{Visual Features}} & \multicolumn{2}{c}{\textbf{64 Compressed Encodings}} & \multicolumn{2}{c}{\textbf{OCR Encodings}} \\
& DocVQA & InfoVQA & DocVQA & InfoVQA\\
\midrule
\rowcolor{cyan!10} $\times$ & 85.5 & 56.8 & 89.2 & 62.9 \\
\rowcolor{cyan!10} $\checkmark$ & 90.2 & 60.2 & 91.9 & 65.3\\ 
\hdashline 
$\boldsymbol{\Delta}$ & {\color{OliveGreen} \textbf{+4.7}} & {\color{OliveGreen} \textbf{+3.4 }} & {\color{OliveGreen} \textbf{+2.7}} & {\color{OliveGreen} \textbf{+2.4}}\\
\bottomrule
\end{tabular}
}
\caption{\textbf{Contribution of Visual Features in DocVLM.} Comparison of DocVLM's performance in text-only mode (without visual features) versus full multimodal operation, using both compressed (64 tokens) and full OCR encodings. Results highlight the complementary benefits of visual information in DocVLM's architecture.} \label{tab: visual features effect}
\vspace{-0.4cm}
\end{table}

\subsection{Exploring LLM Fine-tuning for Text-Only}

\paragraph{Impact of LLM Fine-tuning with LoRA.} To assess the potential for further improvement in DocVLM's text processing capabilities, we fine-tuned the LLM for an additional 100K steps using LoRA, focusing on the text-only mode of operation. \cref{tab: DocVLM_lora_in_llm} presents the results of this experiment, including a comparison with the baseline of inputting OCR words directly.
Our results show that LoRA significantly improves the OCR words baseline performance. However, both compressed and full OCR encodings outperform this improved baseline even without LoRA fine-tuning. Notably, we observed only minor performance improvements when applying LoRA to the LLM with our OCR encodings, both compressed and full.
Based on these findings in the text-only scenario, we decided against additional fine-tuning in our full multimodal DocVLM method. This decision helps maintain the vision and LLM alignment achieved through the extensive pretraining of the original VLM, ensuring that DocVLM enhances the existing VLM abilities without disrupting its pretrained knowledge.
\vspace{-0.2cm}
\begin{table}[h]
\centering
\resizebox{\linewidth}{!}{
\begin{tabular}{c c c c c c c} 
\toprule 
\multirow{2}{*}{\textbf{LoRA}} & \multicolumn{2}{c}{\textbf{OCR Words}} & \multicolumn{2}{c}{\textbf{64 Compressed Encodings}} & \multicolumn{2}{c}{\textbf{OCR Encodings}} \\
& DocVQA & InfoVQA & DocVQA & InfoVQA & DocVQA & InfoVQA\\
\midrule
\rowcolor{cyan!10} $\times$ & 76.4 & 44.5 & 85.5 & 56.8 & 89.2 & 62.9 \\
\rowcolor{cyan!10} $\checkmark$ & 80.3 & 49 & 85.7 & 56.8 & 89.4 & 63\\ 
\hdashline 
$\boldsymbol{\Delta}$ & {\color{OliveGreen} \textbf{+3.9}}& {\color{OliveGreen} \textbf{+4.5}}
& {\color{OliveGreen} \textbf{+0.2}} & {\color{OliveGreen} \textbf{+0 }} & {\color{OliveGreen} \textbf{+0.2}} & {\color{OliveGreen} \textbf{+0.1}}\\
\bottomrule
\end{tabular}
}
\caption{\textbf{Effect of LoRA Fine-tuning on Text-Only Performance.} Comparison before and after LoRA fine-tuning in text-only mode for OCR words baseline, compressed and full OCR encodings. Results show minimal gains for DocVLM's encodings.} \label{tab: DocVLM_lora_in_llm}
\vspace{-0.2cm}
\end{table}


\end{document}